\documentclass[journal]{IEEEtran}

\usepackage{times}
\usepackage{epsfig}
\usepackage{graphicx}
\usepackage{amsmath}
\usepackage{amssymb}
\usepackage{enumitem}
\usepackage{array}
\usepackage{multirow}
\usepackage{graphicx}
\usepackage{subfigure}
\usepackage{amsthm,amsmath,amssymb}
\usepackage{mathrsfs}
\usepackage{booktabs}
\usepackage{cite}
\usepackage{color}
\usepackage[ruled,vlined,linesnumbered]{algorithm2e}

\ifCLASSINFOpdf
\else
\fi

\begin{document}
\title{A Cooperation-Aware Lane Change Method for Autonomous Vehicles}

\author{\IEEEauthorblockN{Zihao Sheng,
Lin Liu,
Shibei Xue, \IEEEmembership{Senior Member, IEEE},
Dezong Zhao, \IEEEmembership{Senior Member, IEEE},\\
Min Jiang,
Dewei Li}
\thanks{This work was supported in part by the National Natural Science Foundation of China (NSFC) under Grants 61873162, 61973317, and in part by the Open Research Project of the State Key Laboratory of Industrial Control Technology, Zhejiang University, China (No. ICT2021B24), and in part by the Scientific Research Funding of Shanghai Dianji University (No. B1-0288-21-007-01-023), and in part by the Engineering and Physical Sciences Research Council of the U.K., through the EPSRC Innovation Fellowship, under Grant EP/S001956/1.({\it Corresponding author: Shibei Xue.})}
\thanks{Zihao Sheng, Shibei Xue and Dewei Li are with the Department of Automation, Shanghai Jiao Tong University and the Key Laboratory of System Control and Information Processing, Ministry of Education of China, Shanghai, 200240 (E-mails: \{zihaosheng, shbxue,dwli\}@sjtu.edu.cn).
Lin Liu is with the School of Electronic Information Engineering, Shanghai Dianji University, Shanghai 201306, People’s Republic of China (E-mail: liul@sdju.edu.cn).
Dezong Zhao is with the James Watt School of Engineering, University of Glasgow, Glasgow, G12 8QQ, United Kingdom  (E-mail: Dezong.Zhao@glasgow.ac.uk).
Min Jiang is with the School of Electronics and Information Engineering, Soochow University, Suzhou 215006, People’s Republic of China (E-mail: jiangmin08@suda.edu.cn)
}}

\markboth{IEEE TRANSACTIONS ON VEHICULAR TECHNOLOGY}%
{Sheng \MakeLowercase{\textit{et al.}}: Title of this paper XXX.}

\maketitle

\begin{abstract}

Lane change for autonomous vehicles (AVs) is an important but challenging task in complex dynamic traffic environments.
Due to difficulties in guarantee safety as well as a high efficiency, AVs are inclined to choose relatively conservative strategies for lane change.
To avoid the conservatism, this paper presents a cooperation-aware lane change method utilizing interactions between vehicles. We first propose an interactive trajectory prediction method to explore possible cooperations between an AV and the others.
Further, an evaluation is designed to make a decision on lane change, in which safety, efficiency and comfort are taken into consideration.
Thereafter, we propose a motion planning algorithm based on model predictive control (MPC), which incorporates AV’s decision and surrounding vehicles’ interactive behaviors into constraints so as to avoid collisions during lane change.
Quantitative testing results show that compared with the methods without an interactive prediction, our method enhances driving efficiencies of the AV and other vehicles by 14.8\% and 2.6\% respectively, which indicates that a proper utilization of vehicle interactions can effectively reduce the conservatism of the AV and promote the cooperation between the AV and others.

\end{abstract}


\IEEEpeerreviewmaketitle

\section{Introduction}
\IEEEPARstart{A}{utonomous} driving technologies are believed to have potentials to significantly reduce traffic accidents, improve travel efficiencies, and enhance comforts of drivers and passengers \cite{paden2016survey,rios2016survey}.
To this end, AVs should have abilities to complete different tasks, such as cruise, lane change, etc. Lane change therein is regarded as one of the most dangerous tasks, due to interactions with surrounding vehicles, and causes many traffic accidents.
However, lane change would benefit to greatly save travelling times and ensure safety, when it is performed at a right time and in a proper way.
In order to enhance the performance of autonomous driving, many studies focus on decision making and motion planning associated with lane change.

In the aspect of decision making on lane change, rule-based approaches were widely used in previous works since they are clear in logic and with low computational costs.
For example, Wei {\it et al.} \cite{wei2010prediction} proposed a decision making framework to determine a target lane and an acceleration, where a distance-keeping model predicts other vehicles’ trajectories and a cost function was designed for selection of the best decision.
Moridpour {\it et al.} \cite{moridpour2012lane} proposed a fuzzy logic model for decision in lane change, in which a fuzzy rule set was defined to explain decision making processes.
However, rule-based methods lack flexibility and only handle limited scenarios because of the difficulty in the logic design for complex or unknown scenarios.
In addition, it is difficult for rule-based methods to handle uncertainties.
To outperform it, Markov decision process (MDP) \cite{ulbrich2013probabilistic,coskun2018predictive,sezer2018intelligent} has been studied to provide robust lane change decisions.
However, it is generally difficult to solve MDP in real time due to its high computational complexity \cite{schwarting2018planning}.

In recent years, with the development of machine learning, learning-based methods provide a new avenue for lane change decisions. Machine learning models not only can deal with uncertainties due to their probabilistic reasoning, but also can learn nonlinear mapping relations from real-world data to decisions, such that they improve the robustness of decisions.
In this way, learning-based methods can make human-like decisions.
For example, Hou {\it et al.} \cite{hou2013modeling} developed a lane-change-assisted framework, which combines a decision tree and a Bayes classifier to predict whether other drivers decide to merge.
Gu {\it et al.} \cite{gu2020novel} used a deep autoencoder network to identify driving behaviors and a XGBoost model to decide whether to change lanes.
Other learning-based approaches were also proposed for lane change decisions including support vector machine \cite{zhang2018study,liu2019novel}, deep neural network \cite{zheng2014predicting,costilla2017deep,li2018humanlike} and reinforcement learning \cite{ngai2011multiple,xu2018reinforcement,you2018highway}.
However, due to the lack of interpretability on machine learning algorithms, it is difficult to find the reason of a failure decision under some scenarios.

On the other hand, to complete lane change, an AV with a decision needs to plan a proper trajectory that satisfies specific constraints, e.g., vehicle dynamics, collision avoidance, and available times.
In the existing works, motion planning methods can be generally classified into four categories: graph search, interpolating curve, sampling, and numerical optimization \cite{gonzalez2015review}.
The commonly used graph-search-based methods include $D^*$ algorithm \cite{stentz1997optimal}, field $D^*$ algorithm \cite{ferguson2006using}, Dijkstra algorithm \cite{parungao2018dijkstra}, and hybrid $A^*$ algorithm \cite{fassbender2014trajectory}.
However, these methods rarely consider constraints of vehicle kinematics, so the generated paths are usually difficult to track.
In addition, interpolating-curve-based approaches are usually efficient in computation, such as polynomial curves \cite{guan2005robotic,he2018human}, Bezier curves \cite{han2010bezier,qian2016motion}, Dubins curves \cite{yang20132d}, and Clothoids curves \cite{bertolazzi2018efficient}.
Nevertheless, these curves are sequences of positions without considering the variation of positions over time, so they cannot guarantee to avoid collisions.
Moreover, sampling-based methods sample a configuration space and select samples satisfying predefined requirements as results.
Many sampling-based methods were developed based on rapidly-exploring random trees (RRT), since it can deal with general dynamic models \cite{kuffner2000rrt,melchior2007particle,kothari2013probabilistically}.
However, their results would be unstable.
Finally, numerical-optimization-based approaches obtain an optimal trajectory by minimizing an objective function subject to a set of constraints \cite{ziegler2014making,xu2012real}.
This kind of method can comprehensively incorporate various objectives and constraints in trajectory planning.

Although the above approaches have made great progresses in improving AVs’ capacities to execute lane change, there are still limitations.
Firstly, these methods take the decision making and trajectory planning for lane change as two separate problems.
However, they are actually highly interdependent, so the separation of them would lead to inconsistency between each other. For instance, if the mechanical and physical limits of vehicle motions are not considered by decision making, the motion planning module may fail to find a feasible solution.
Secondly, potential cooperations between an AV and other vehicles have not been fully explored, resulting in conservative behaviors of AVs in some complex scenarios. As a specific example, in dense traffic, an AV would not change lanes if it cannot cooperate with others.

Therefore, to overcome the above limitations and improve the safety and efficiency of AVs, this paper presents a cooperation-aware method for determining a proper time instant and a maneuver in lane change, which takes interactions among vehicles into account.
Instead of separately designing the modules of decision making and motion planning, we consider these two modules in a unified framework.
In detail, we first construct a decision set, in which each decision is generated by a positive and negative trapezoidal lateral acceleration (PNTLA) method.
Here, the PNTLA utilizes mechanical and physical limits of vehicle motions, so the feasibility of motion planning is considered during decision making.
In addition, we propose an interactive trajectory prediction method to predict surrounding vehicles’ reactions to each decision, which allows to explore possible cooperations.
Based on these predictions, an optimal decision is selected by considering safeties, efficiencies and comforts.
Subsequently, an MPC-based motion planning algorithm is proposed to solve a trajectory for lane change.
Here, we incorporate the prediction of interactive trajectories into constraints to effectively use vehicular interactions to plan a safe trajectory.
The contributions of this paper are two folds.

(1) Different from existing works, we propose a complete lane change method, which considers the correlation between decision making and trajectory planning to enhance efficiencies for lane change and prevent conflicts between them.
In this way, this method avoids the situations that the planning module fails to generate a trajectory due to an improper decision, and thus guarantees a smooth driving.

(2) Compared with existing methods, we propose an interactive trajectory prediction method, which helps an AV make proactive decisions and avoid being shortsighted.
The comparison with other methods demonstrates that it improves driving efficiencies of the AV and others by 14.8\% and 2.6\%, demonstrating that the AV can use interactions to cooperate with others, and thus to achieve an efficient driving.

The rest of this paper is organized as follows. The problem description is given in Section II. Our system structure is described in Section III. We present a detailed description of the lane change decision module in Section IV. The lane change trajectory planning module is introduced in Section V. Experimental results and analysis are shown in Section VI. Finally, Section VII concludes this paper.

\section{Problem description of lane change}

Lane change is a key maneuver for autonomous driving, which can help an AV secure a higher driving efficiency and safety.
In this paper, we consider that only one vehicle is controlled by our proposed lane change method, referred as “ego vehicle”, and all vehicles are not connected through vehicle-to-vehicle or vehicle-to-infrastructure. All vehicles are passenger cars with similar shapes and motion patterns; motorcycles, buses or heavy trucks are not considered. Since a lane change is not recommended near intersections or traffic signals, we assume that the road remains straight along a lane change.

Lane change is executed in the form of a trajectory, which is a function of time mapping time to the state of a vehicle.
Hence, we define a trajectory of an ego vehicle as $\tau : \mathbb{R} \rightarrow \mathbb{R}^{C}$. For the convenience of trajectory planning, the resulting state is denoted as,
\begin{equation}
\tau(t) = [x(t),~y(t),~\psi(t),~v(t)]^\top,
\end{equation}
where $x$ and $y$ denote longitudinal and lateral positions of a vehicle, respectively, $\psi$ represents the heading angle, and $v$ denotes the velocity.
Here, note that these four components are studied for lane change and thus we have $C=4$.
When a vehicle runs on a road, the drivable areas impose boundaries on $x$ and $y$. Besides, $v$ and $\psi$ are limited by the motion capacity of a vehicle.

When the ego vehicle runs on a road with at least two lanes, we assume that it observes perfect trajectories of $N$ vehicles around it without any noise or delay.
Since the collected data is processed in a digital way, trajectories are discretized with a sampling period $\Delta t$. The discrete trajectory is denoted as $\tau[k] = \tau(k\Delta t)$.
With an observation horizon $T_o$, the observed trajectory of the $n$th vehicle is denoted as,
\begin{equation}
\tau_n^o = \tau_n[k] = [x_n[k],~y_n[k],~\psi_n[k],~v_n[k]]^\top,1 \leq k \leq O,
\end{equation}
where the subscript $n$ indicates the trajectory of the $n$th vehicle, $O = T_o/\Delta t$, $\tau_n[k]$ is the state of the $n$th vehicle at the time step $k$. Similarly, the trajectory of the ego vehicle over $T_o$ is denoted as,
\begin{equation}
\begin{split}
\tau_{ego}^o & = \tau_{ego}[k] = [x_{ego}[k],~y_{ego}[k],~\psi_{ego}[k],~v_{ego}[k]]^\top,\\
& 1 \leq k \leq O.
\end{split}
\end{equation}
After the observation is obtained, the ego vehicle selects an optimal decision on lane change from a decision set $\mathcal{D}$.
The decision set $\mathcal{D}$ consists of $B$ candidate decisions, which is denoted as $\mathcal{D} = \{\tau_i^d ~ | ~ \forall i \in \{1,\cdots, B\}\}$. Each decision corresponds to a reference trajectory, denoted as,
\begin{equation}
\tau_b^d = \tau_b[k] = [x_b[k],~y_b[k],~\psi_b[k],~v_b[k]]^\top,O+1 \leq k \leq O+D,
\end{equation}
with a duration $T_d$, where the subscript $b$ denotes the index of a decision, $D = T_d/\Delta t$, $\tau_b[k]$ is the state of the $b$th decision at the time step $k$.
The optimal decision on lane change is denoted as $\tau_{opt}^d$.
Based on $\tau_{opt}^d$, the ego vehicle adjusts its acceleration and steering angle to plan a trajectory over a planning horizon $T_p$, which is represented as,
\begin{equation}
\tau_{ego}^p = \tau_{ego}[k], ~ O+1\leq k \leq O+P,
\end{equation}
where $P = T_p / \Delta t$.

Thus, the problem of lane change is summarized as follows.
Given observed trajectories of the ego vehicle and its surrounding vehicles, i.e., $\tau_{ego}^o$ and $\{\tau_1^o, \cdots, \tau_n^o, \cdots, \tau_N^o\}$, the ego vehicle selects an optimal decision $\tau_{opt}^d$ from a decision set $\mathcal{D}$, and finally plans a trajectory $\tau_{ego}^p$ by adjusting its acceleration and steering angle.

\begin{figure}
\centerline{\includegraphics[scale=0.3]{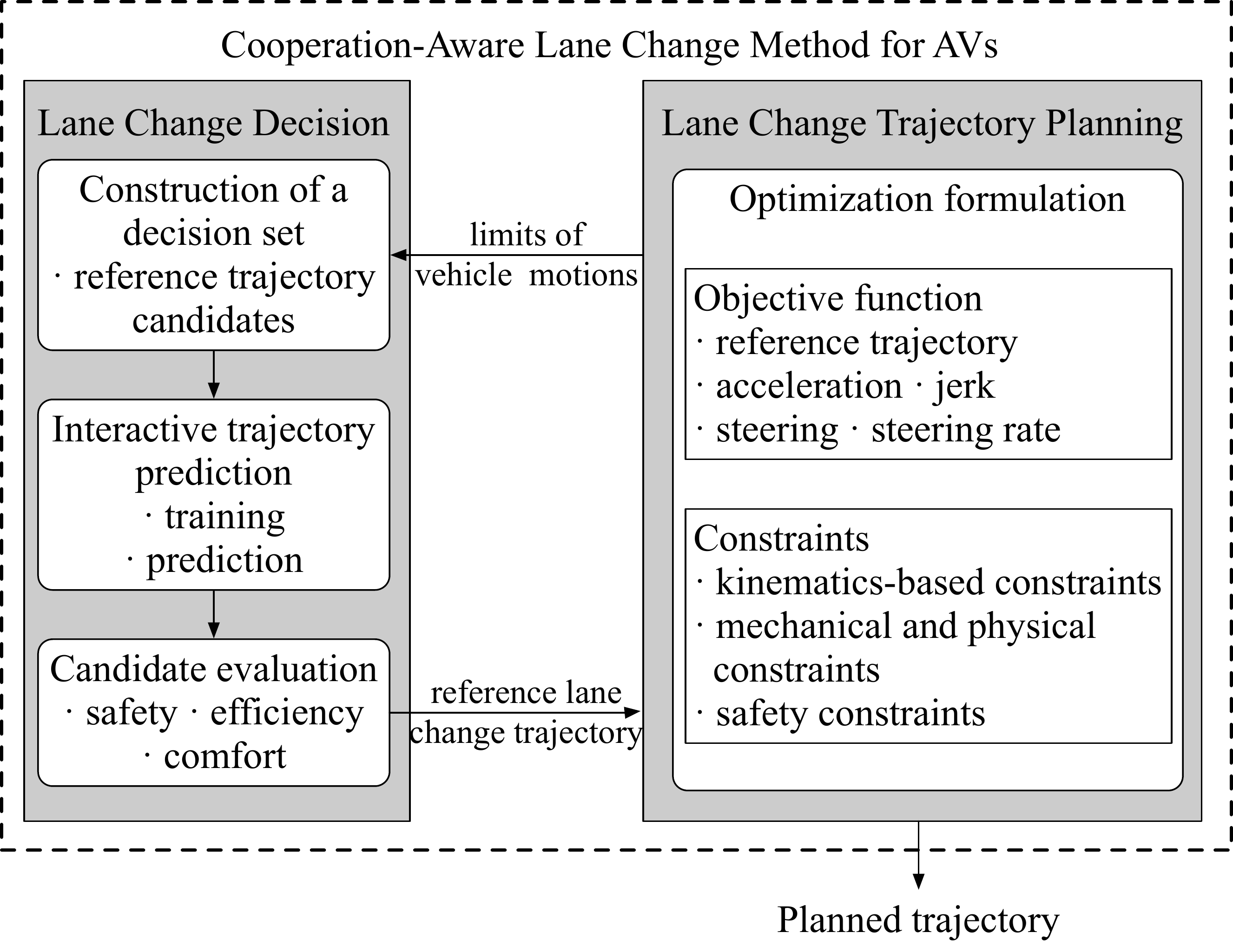}}
\caption{System architecture of the proposed cooperation-aware lane change method for autonomous vehicles.}
\label{frame}
\end{figure}

\section{Structure of proposed cooperation-aware lane change method}
As mentioned before, most studies on lane change treat decision making and trajectory planning as two separate problems. The decision making module provides high-level orders, such as accelerating/decelerating, left/right lane change, to low-level modules. The trajectory planning module receives these orders and then solves a feasible trajectory. Although decision making and trajectory planning are hierarchical in term of modules, they are actually highly coupled for successful executions \cite{hang2020integrated,lu2020hierarchical}. Therefore, the separate design of the decision making and trajectory planning causes that an improper decision from the decision making module is sent to the trajectory planning module, which might fail to generate a feasible trajectory.

To solve the above problem, we propose a cooperation-aware lane change method, which considers the correlation between the decision making and trajectory planning.
The structure of the proposed method is shown in Fig. \ref{frame}.
We can see that the two modules are interdependent.
The planning module first sends limits of vehicle motions to the decision module, so that the feasibility of trajectory planning can be considered in decision making.
Then, the decision module provides a reference trajectory for the planning module, which uses it to generate a planned trajectory.

The main aim of the decision module is to determine an appropriate time and a proper speed to change lanes while considering the limits of vehicle motions.
We first generate various reference trajectory candidates, which consider the feasibility of trajectory planning, to construct a decision set $\mathcal{D}$.
An interactive trajectory prediction method is then proposed to explore which reference trajectory candidate would potentially promote cooperations between the ego vehicle and surrounding vehicles.
Finally, an optimal decision $\tau_{opt}^d$ is selected from $\mathcal{D}$ based on the interactive trajectory prediction while considering safety, driving efficiency and comfort.

The role of the trajectory planning module is to generate a dynamically feasible, collision-free and smooth trajectory based on $\tau_{opt}^d$.
The optimal decision $\tau_{opt}^d$ poses constraints for the trajectory planning.
We formulate the trajectory planning as an optimization problem, which takes model predictive control (MPC) as the basis, since MPC can comprehensively consider the objective function and various constraints, including vehicle dynamics, safety, smooth control actions, ect.
By solving this optimization problem, a qualified trajectory is planned.

\section{Lane change decision based on interactive trajectory prediction}\label{section-decision}

In this section, we introduce details about the decision module for lane change in the left side of Fig. \ref{frame}, including the construction of decision set, interactive trajectory prediction, and candidate evaluation.

\subsection{Construction of a decision set}\label{generate}

\begin{figure}
\centerline{\includegraphics[scale=0.4]{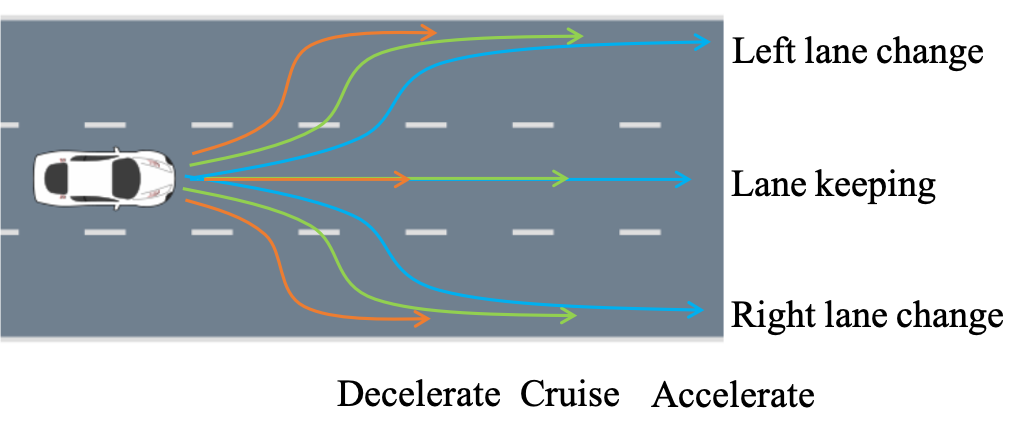}}
\caption{Predefined reference trajectory candidates which combine three longitudinal options and three lateral actions.}
\label{decision}
\end{figure}

The decision set $\mathcal{D}$ consists of various candidate decisions. Each decision $\tau_b^d$ is a reference trajectory, which is an initialization for the subsequent interactive prediction and trajectory planning.
For the computational convenience, we set the duration of each reference trajectory $T_d$ to be equal to $T_p$.
In this paper, we limit our discussion on combining three lateral actions (left lane change, right lane change and staying in the current lane) and three longitudinal options (accelerating, decelerating, and keeping in the current speed range) in Fig. \ref{decision} to construct $\mathcal{D}$. Note that $\mathcal{D}$ can be extended to contain more decisions according to a practical demand.

\begin{figure}
\centerline{\includegraphics[scale=0.45]{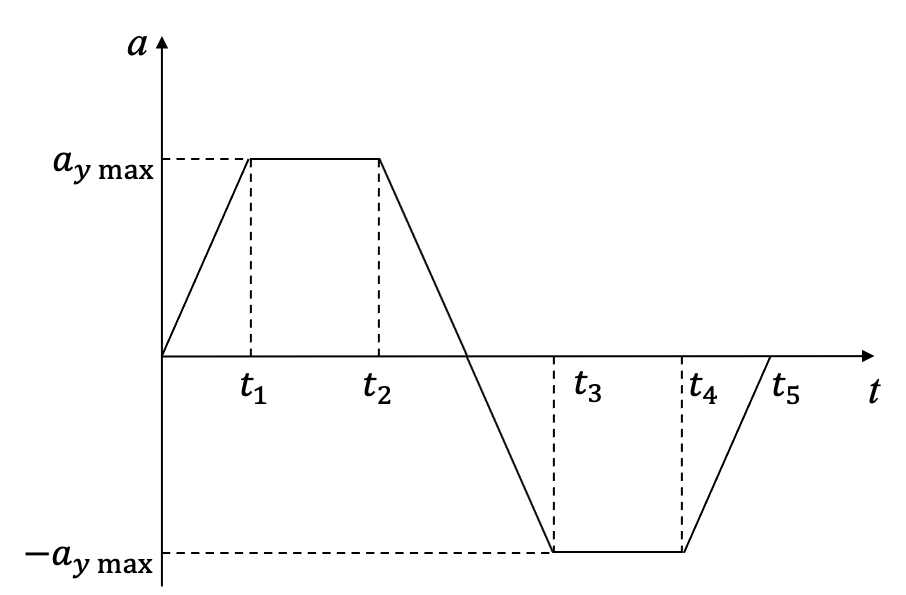}}
\caption{Curve of lateral acceleration generated by the positive and negative trapezoidal lateral acceleration method.}
\label{trapezoid}
\end{figure}

In order to generate each $\tau_b^d$, we first calculate the lateral and the longitudinal accelerations.
Inspired by the work in \cite{guo2014lane}, we use the PNTLA method to generate a continuous-time lateral reference trajectory. This method can calculate the smooth trajectory with curvature continuity based on the maximum lateral acceleration $a_{y\max}$ and the maximum lateral jerk $\Delta a_{y\max}$.
As shown in Fig. \ref{trapezoid}, the PNTLA method assumes that when vehicles change lanes, the lateral acceleration varies linearly, and the acceleration curve is composed of two isosceles trapezoids with the same size but opposite directions.
Hence, the lateral jerk $\Delta a_y$ is expressed as,
\begin{equation}
\Delta a_y(t)=\left\{
\begin{array}{crr}
\Delta a_{y\max} & & 0 < t < t_1\\
0 & & t_1 \leq t < t_2\\
-\Delta a_{y\max} & & t_2 \leq t < t_3\\
0 & & t_4 \leq t < t_4\\
\Delta a_{y\max} & & t_4 \leq t < t_5\\
\end{array} \right.,
\end{equation}
where $t_1$ to $t_5$ are five key timestamps. Based on the properties of isosceles trapezoids, they are calculated as \cite{guo2014lane},
\begin{equation}
\left\{
\begin{aligned}
t_1 & = \frac{a_{y\max}}{\Delta a_{y\max}}, \\
t_2 & = -\frac{t_1}{2} + \frac{\sqrt{t_1^2+4d_w/a_{y\max}}}{2}, \\
t_3 & = 2t_1 + t_2, \\
t_4 & = t_1 + 2t_2, \\
t_5 & = 2t_1 + 2t_2, \\
\end{aligned}
\right.
\end{equation}
where $d_w$ denotes the lateral displacement during a lane change.
With the timestamps $t_1$ to $t_5$, the lateral acceleration curve can be determined.

The longitudinal reference trajectory is generated in a similar manner. The longitudinal acceleration curve is described as an isosceles trapezoid which guarantees the smoothness and curvature continuity of a generated trajectory.
The longitudinal jerk $\Delta a_x$ is expressed as,
\begin{equation}
\Delta a_x(t)=\left\{
\begin{array}{crr}
\Delta a_{x\max} & & 0 < t < t_6\\
0 & & t_6 \leq t < t_7\\
-\Delta a_{x\max} & & t_7 \leq t < t_8\\
\end{array} \right.,
\end{equation}
where  $\Delta a_{x\max}$ denotes a maximum longitudinal jerk, $t_6$ to $t_8$ are three key timestamps. Based on the properties of isosceles trapezoids, $t_6$ to $t_8$ are calculated as,
\begin{equation}
\left\{
\begin{aligned}
t_6 & = \frac{a_{x\max}}{\Delta a_{x\max}}, \\
t_7 & = \frac{|v_{x1} - v_{x0}| }{a_{x\max}}, \\
t_8 & = t_6 + t_7, \\
\end{aligned}
\right.
\end{equation}
where  $a_{x\max}$ denotes a maximum longitudinal acceleration, $v_{x0}$ is the current speed, $v_{x1}$ is the desired speed.
Similarly, with the timestamps $t_6$ to $t_8$, the longitudinal acceleration curve is established.

After obtaining the lateral and longitudinal acceleration curves, we can calculate a continuous-time velocity profile by integrating the acceleration curves, and then calculate a continuous-time position profile by integrating the acceleration curves twice. Since the reference trajectory does not consider the vehicle dynamics, the heading angle is set to be equal to the angle between the direction of velocity and the longitudinal axis. Therefore, we obtain a continuous-time trajectory, which should be discretized with $\Delta t$ to obtain each $\tau_b^d$.

\subsection{Interactive trajectory prediction of surrounding vehicles}\label{itp}

\begin{figure}
\centerline{\includegraphics[scale=0.36]{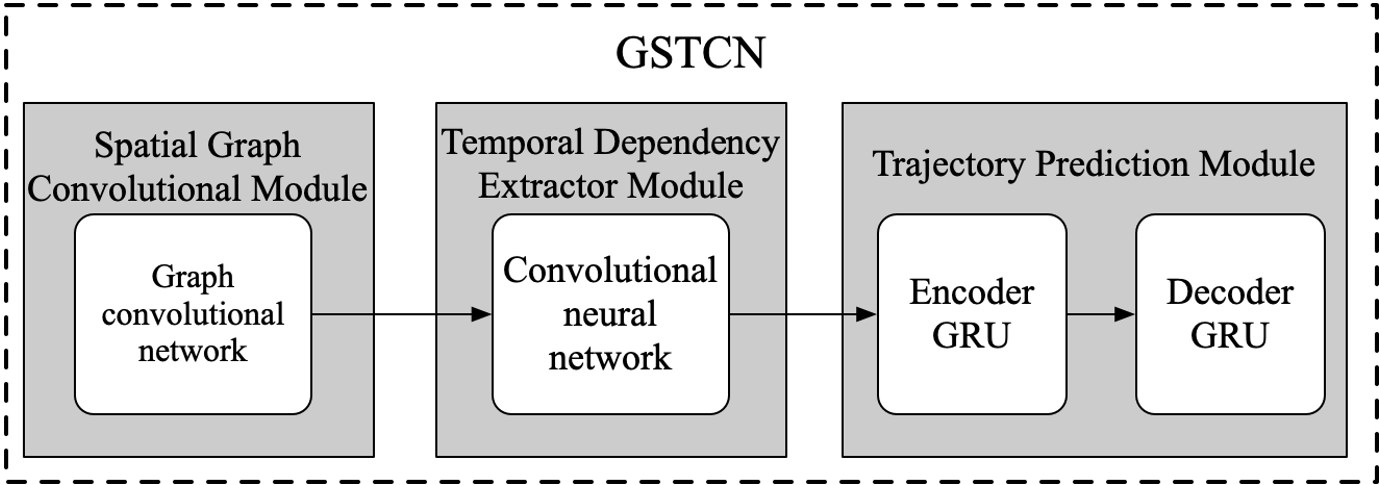}}
\caption{Architecture of GSTCN \cite{sheng2021graph}.}
\label{gstcn}
\end{figure}

To select a $\tau_b^d \in \mathcal{D}$ as an optimal decision $\tau_{opt}^d$ for the ego vehicle, it is crucial to predict how surrounding vehicles will react to each $\tau_b^d$.
However, due to the strong spatial-temporal dependencies among vehicles, predicting their trajectories is challenging. In this paper, we propose an interactive trajectory prediction method based on the graph-based spatial-temporal convolutional network (GSTCN) \cite{sheng2021graph} to tackle this task, since this network can effectively capture the spatial-temporal dependencies among vehicles and then predict their future trajectories simultaneously.
As shown in Fig. \ref{gstcn}, GSTCN is composed of three modules: a spatial graph convolutional module, a temporal dependency extractor module, and a trajectory prediction module. The first module uses a graph convolution network to extract the spatial dependencies between vehicles. The following temporal dependency extractor module learns the temporal dependencies. The trajectory prediction module consists of an encoder and a decoder, where both of them are composed of the gated recurrent unit (GRU) networks to deal with the trajectory sequence generation.
More details of the GSTCN can be found in \cite{sheng2021graph}.
Here, the proposed interactive trajectory prediction method will be introduced in detail from two stages: training step and prediction step.

\subsubsection{Training step} GSTCN takes the observed trajectories of all vehicles over an observation horizon $T_o$ as inputs and predicts their future coordinates simultaneously. GSTCN assumes future positions of vehicles are random variables satisfying bi-variable Gaussian distributions, so this network is trained by minimizing a negative log-likelihood loss. The training step is done offline.

\subsubsection{Prediction step} After the training is done, we use GSTCN to predict surrounding vehicles’ trajectories for each $\tau_b^d$. For a new observation of the positions of surrounding vehicles at the current moment, GSTCN first predicts their positions at the next time step. Then, we use this prediction to update their current positions.
Similarly, the current position of the ego vehicle is updated by $\tau_b^d$.
This process will be repeated until $\tau_b^d$ is used up.
In this way, we obtain the future trajectory $\tau_n^f$ of each surrounding vehicle to each $\tau_b^d$ over a planning horizon $T_p$.

To formalize, the proposed interactive trajectory prediction method is denoted as a function $\phi$ that maps $\tau_b^d$, $\tau_{ego}^o$, and $\{\tau_1^o, \cdots, \tau_n^o, \cdots, \tau_N^o\}$ to predicted interactive trajectories $\{\tau_{1,b}^f, \cdots, \tau_{n,b}^f, \cdots, \tau_{N,b}^f\}$.

\subsection{Candidate evaluation method}
After obtaining the predicted interactive trajectories $\{\tau_{1,b}^f, \cdots, \tau_{n,b}^f, \cdots, \tau_{N,b}^f\}$ under every $\tau_b^d \in \mathcal{D}$, we should determine which $\tau_b^d$ is the most recommended in terms of safety, efficiency and comfort \cite{hang2020human,hang2020integrated}.
We evaluate each $\tau_b^d$ by calculating their costs.
The cost function consists of three components and is calculated as
\begin{equation}
J_d(\tau_b^d) =\lambda_sJ_s + \lambda_eJ_e + \lambda_cJ_c, ~ \tau_b^d \in \mathcal{D},
\end{equation}
where $J_s$, $J_e$, and $J_c$ are the costs of safety, efficiency and comfort, respectively. $\lambda_s$, $\lambda_e$, and $\lambda_c$ represent the corresponding weights, which can be customized by the user.

The safety cost $J_s$ is related to the collision risks between the ego vehicle and nearby vehicles, which consists of risks in both lateral and longitudinal directions.
Therefore, the safety cost $J_s$ is represented as,
\begin{equation}
J_s = J_{s-lon} + \eta^2 J_{s-lat},
\end{equation}
where $J_{s-lon}$ and $J_{s-lat}$ represent the safety costs in longitudinal and lateral directions, respectively. $\eta=1$ if the ego vehicle changes to other lanes, otherwise $\eta=0$, which guarantees that if the $\tau_b^d$ does not relate to lane change, the lateral risks can be ignored.

The longitudinal safety cost $J_{s-lon}$ relates to the collision risks between the ego vehicle and the preceding and following vehicles in the same lane. In this paper, we use the time-to-collision (TTC) \cite{hayward1972near} to characterize the collision risks, which is calculated based on the relative distances and velocities against the ego vehicle and the preceding and following vehicles in the same lane.
The smaller TTC is, the larger the collision risk is.
Thus, we define $J_{s-lon}$ as,
\begin{equation}
J_{s-lon} = \kappa_{s-lon}\sum_{k=O+1}^{O+P}[\sigma(\frac{\Delta v_{pv-lon}[k]}{\Delta s_{pv-lon}[k] }) + \sigma(\frac{\Delta v_{fv-lon}[k]}{\Delta s_{fv-lon}[k] })],
\end{equation}
with
\begin{equation}
\left\{
\begin{aligned}
\Delta v_{pv-lon}[k] & = v_{ego}[k] - v_{pv-lon}[k], \\
\Delta v_{fv-lon}[k] & = v_{ego}[k] - v_{fv-lon}[k], \\
\Delta s_{pv-lon}[k] & = \left\| s_{ego}[k] - s_{pv-lon}[k] \right\|_2, \\
\Delta s_{fv-lon}[k] & = \left\| s_{ego}[k] - s_{fv-lon}[k] \right\|_2, \\
\end{aligned}
\right.
\end{equation}
where $v_{ego}[k]$ obtained from $\tau_b^d$ is the velocity of the ego vehicle at the time step $k$.
The subscripts $pv-lon$ and $fv-lon$ represent the indices of the nearest preceding and following vehicles in the same lane, respectively. $v_{pv-lon}$ is the velocity of the nearest preceding vehicle in the same lane at the time step $k$, which is obtained from $\{\tau_{1,b}^f, \cdots, \tau_{n,b}^f, \cdots, \tau_{N,b}^f\}$. $s=[x,y]^{\top}$ denotes the position of a vehicle, and $\left\| \cdot \right\|_2 $ is the $l_2$ norm.
Hence, $\Delta v_{pv-lon}[k]$ and $\Delta s_{pv-lon}[k]$ denote the relative velocity and distance between the ego vehicle and the nearest preceding vehicle in the same lane at the time step $k$.
$\Delta v_{fv-lon}$ and $\Delta s_{fv-lon}$ denote the relative velocity and distance between the ego vehicle and the nearest following vehicle in the same lane.
$\kappa_{s-lat}$ is the regularization coefficient of the longitudinal safety cost, which is determined by practical demands. $\sigma(\cdot)=\min(\cdot, 0)$, which guarantees that two vehicles moving away from each other will not contribute a safety cost.

Similarly, the lateral safety cost $J_{s-lat}$ is calculated based on the relative distances and velocities against the ego vehicle and the preceding and following vehicles in the target lane, which is defined as,
\begin{equation}
J_{s-lat} = \kappa_{s-lat}\sum_{k=O+1}^{O+P}[\sigma(\frac{\Delta v_{pv-lat}[k]}{\Delta s_{pv-lat}[k] }) + \sigma(\frac{\Delta v_{fv-lat}[k]}{\Delta s_{fv-lat}[k] })],
\end{equation}
where $\kappa_{s-lat}$ is the regularization coefficient of the lateral safety cost.
The subscripts $pv-lat$ and $fv-lat$ represent the indices of the nearest preceding and following vehicles in the target lane, respectively.

Moreover, the efficiency cost $J_e$ is related to the velocities of the ego vehicle and surrounding vehicles.
The closer their speeds are to the desired speed, the smaller the efficiency cost is.
Hence, $J_e$ is defined as,
\begin{equation}
J_e = \kappa_e\sum_{k=O+1}^{O+P}(v_{ego}[k]-v_{des})^2+\frac{\kappa_e}{N}\sum_{n=1}^{N}\sum_{k=O+1}^{O+P}(v_{n}[k]-v_{des})^2,
\end{equation}
where $\kappa_{e}$ is the regularization coefficient of the efficiency cost, $v_{des}$ is the desirable velocity, $v_n$ and $v_{ego}$ are the velocities of the $n$th nearby vehicle and the ego vehicle, respectively.
Since we want to make decisions that can help the ego vehicle to cooperate with others, the efficiency cost indicates that both the ego vehicle and surrounding vehicles want to reach the desirable speeds.

In addition, the comfort cost $J_c$ aims to make the ego vehicle travel with a relatively smooth trajectory, so it relates to the jerk in both lateral and longitudinal directions.
$J_c$ is defined as,
\begin{equation}
J_c = \sum_{k=O+1}^{O+P}[\kappa_{c-lon}(\Delta a_{ego-lon}[k])^2 + \kappa_{c-lat}(\Delta a_{ego-lat}[k])^2],
\end{equation}
where $\kappa_{c-lon}$ and $\kappa_{c-lat}$ are regularization coefficients of the comfort cost. $\Delta a_{ego-lon}$ and $\Delta a_{ego-lat}$ denote the longitudinal and lateral jerks of the ego vehicle, which can be obtained by calculating the derivative of acceleration with respect to time from $\tau_b^d$.

After the cost function is defined, we select a $\tau_b^d \in \mathcal{D}$ with the minimal cost $J_d$ as the optimal decision $\tau_{opt}^d$, which makes a balance among safety, efficiency, and comfort.

Note that in the process of generating the reference trajectory candidates, we utilize the lateral displacement, the maximum lateral acceleration, the maximum lateral jerk, the maximum longitudinal acceleration, and the maximum longitudinal jerk.
Therefore, the feasibility of a motion planning has been considered during the phase of the lane change decision.

\section{Model-predictive-control-based trajectory planning for lane change}
In this section, based on the above optimal reference trajectory $\tau_{opt}^d$ and the corresponding interactive trajectory prediction $\{\tau_{1,opt}^f, \cdots, \tau_{n,opt}^f, \cdots, \tau_{N,opt}^f\}$, we introduce the trajectory planning for a lane change.
We formulate the trajectory planning as a model predictive control (MPC)-based optimization problem.
MPC can effectively handle optimization problems with constraints while respecting vehicle dynamics.
Next, we will introduce a predictive model, constraints, an objective function, and the Ipopt-based approach for trajectory planning.

\subsection{Kinematics-based predictive model}
To realize the trajectory planning for the ego vehicle, we first establish a mathematical model of vehicle kinematics. The more accurately the model describes the motions of a vehicle, the more accurately the planned trajectory will be tracked by a control system. However, due to the limited computing resources and high requirements for computing time in driving environments, the vehicle kinematics model should also have the advantages of easy implementation and low computation complexity.
To this end, we use the bicycle model in \cite{kong2015kinematic} to describe the nonlinear vehicle kinematics. With the Euler discretization, this vehicle kinematic model can be defined as,
\begin{equation} \label{vkm}
\left\{
  \begin{aligned}
  x[k+1] &= x[k] + \Delta t \times v[k]\cos(\psi[k] + \beta[k]) ,\\
  y[k+1] &= y[k] + \Delta t \times v[k]\sin(\psi[k] + \beta[k]), \\
  \psi[k+1] &= \psi[k] + \Delta t \times \frac{v[k]}{l_r}\sin(\beta[k]), \\
  v[k+1] &= v[k] + \Delta t \times a[k], \\
  \beta[k] &= \tan^{-1}(\frac{l_r}{l_r+l_f}\tan(\delta[k])),
\end{aligned}
\right.
\end{equation}
where $\beta$ denotes the angle between the speed and the heading, $l_f$ and $l_r$ denote the length from the gravity center of a vehicle to the front axle and the rear axle, respectively, $a$ is the acceleration, and $\delta$ denotes the steering angle.
Thus, we obtain a predictive model for MPC.
As mentioned in Section II, the acceleration and the steering angle are taken as control actions, and thus we define them as $\mathbf{u}[k]=[a[k],~\delta[k]]^\top$.

\subsection{Mechanical and physical constraints}
Due to the mechanical and physical limitations in velocity, heading angle, and position, we need to take the boundaries of these variables as constraints in the process of motion planning to achieve a feasible solution. Specifically, they are represented as,
\begin{equation}\label{pc}
\left\{
\begin{array}{lcl}
\tau_{\min} \leq  \tau[k]  \leq \tau_{\max}, \\
\mathbf{u}_{\min} \leq  \mathbf{u}[k] \leq \mathbf{u}_{\max} ,\\
\Delta\mathbf{u}_{\min} \leq  \mathbf{u}[k]-\mathbf{u}[k-1]  \leq \Delta\mathbf{u}_{\max},
\end{array} \right.
\end{equation}
where the first constraint considers the road boundaries and the available range of the heading angle and the velocity, i.e., $\tau_{\min}$ and $\tau_{\max}$; the second constraint ensures the acceleration and the steering angle do not exceed the vehicle's capacity from $\mathbf{u}_{\min}$ to $\mathbf{u}_{\max}$; the last constraint guarantees the smoothness and comfort of generated trajectories by restraining the control input rate to a range from $ \Delta\mathbf{u}_{\min}$ to $\Delta\mathbf{u}_{\max}$.

\subsection{Safety constraints}
\begin{figure}
\centerline{\includegraphics[scale=0.4]{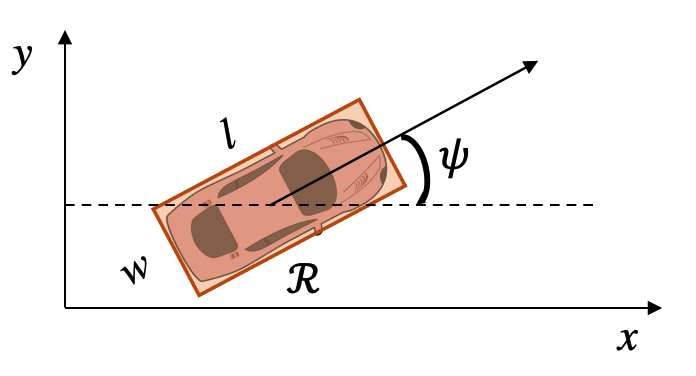}}
\caption{The road area occupied by a vehicle is represented by a rectangle rotated by the heading angle $\psi$ in the two-dimensional plane.}
\label{shape}
\end{figure}

In addition to the above mechanical and physical constraints, we also consider the safety constraints.
Since the vehicle-occupied road area imposes restrictions on the movements of other vehicles, it is essential to model a precise vehicle shape.
Unlike previous works that model the shape of vehicles as a mass point or particles with radial gaps, we assume vehicles are rectangular which is closer to an actual circumstance.
As shown in Fig. \ref{shape}, the vehicle-occupied road region is represented by a rectangle $\mathcal{R}$, which is expressed as,
\begin{equation}
\mathcal{R} = \{\mathbf{p}=[p_1,~p_2]^{\top} |~ \mathbf{A} \cdot \mathbf{p} \leq \mathbf{b} \}.
\end{equation}
As in \cite{firoozi2021formation}, the matrix $\mathbf{A}$ and the vector $\mathbf{b}$ are expressed as,
\begin{equation}\label{occupied-A-b}
\begin{aligned}
\mathbf{A} &= \begin{bmatrix} \quad\cos \psi & \quad\sin \psi \\ -\sin \psi & \quad\cos \psi  \\
                 -\cos \psi & -\sin \psi \\ \quad\sin \psi & -\cos \psi  \end{bmatrix},  \\
\mathbf{b} &= [l/2, w/2, l/2, w/2]^{\top} + \mathbf{A} \cdot [x_0, y_0]^{\top},
\end{aligned}
\end{equation}
with the length $l$ and width $w$ of a vehicle, and the current coordinate of the gravity center $[x_0, y_0]$.

With this vehicle shape model, the distance between two vehicles can be represented as,
\begin{equation}\label{dist_cal}
{\rm dist}(\mathcal{R}_1, \mathcal{R}_2) = \{\min_{\mathbf{p}_1, \mathbf{p}_1} \left\| \mathbf{p}_1 - \mathbf{p}_2 \right\|_2 ~ | ~ \mathbf{p}_1\in \mathcal{R}_1,~\mathbf{p}_2 \in \mathcal{R}_2 \} ,
\end{equation}
where $\mathcal{R}_1$ and $\mathcal{R}_2$ are the road regions occupied by two vehicles, respectively, and $\mathbf{p}_1$ and $\mathbf{p}_2$ denote any coordinates in the occupied road regions.
Therefore, the safety constraints is represented as,
\begin{equation}\label{sc}
{\rm dist}(\mathcal{R}_{ego}[k], \mathcal{R}_n[k]) > d_{\min}, \forall n \in \{1, \cdots, N\},
\end{equation}
where $\mathcal{R}_{ego}[k]$ and $\mathcal{R}_n[k]$ denote the road regions occupied the ego vehicle and the $n$th nearby vehicle at the time step $k$, respectively, $d_{\min}$ is the minimum safety distance between vehicles.
With the interactive trajectory prediction of nearby vehicles $\{\tau_{1,opt}^f, \cdots, \tau_{n,opt}^f, \cdots, \tau_{N,opt}^f\}$, we can calculate their occupied regions in the planning time horizon. In this way, vehicular interactions are considered during trajectory planning.

Note that the collision avoidance constraint (\ref{sc}) is nondifferentiable, so we adopt the approach in \cite{zhang2020optimization} to convert this constraint to a smooth and differentiable constraint by using strong duality theory.
Based on this method, the safety constraints (\ref{sc}) can be reformulated as,
\begin{equation}\label{sc-re}
\left\{
\begin{array}{l}
-\mathbf{b}[k]^{\top}\mathbf{\lambda}_n[k] -\mathbf{b}_n^{\top}[k]\mathbf{\mu}_n[k] \geq d_{\min} \\
\mathbf{A}[k]^{\top}\mathbf{\lambda}_n[k] + \rho_n[k] = 0 \\
\mathbf{A}_n^{\top}[k]\mathbf{\mu}_n[k] - \rho_n[k] = 0 \\
\rho^{\top}_n[k]\rho_n[k] \leq 1  \\
\mathbf{\lambda}_n[k] \geq 0\\
\mathbf{\mu}_n[k] \geq 0
\end{array} \right., \forall n \in \{1, \cdots, N\},
\end{equation}
where $\mathbf{\lambda}_n$, $\mathbf{\mu}_n$ and $\rho_n$ are the dual variables. Interested readers are referred to \cite{zhang2020optimization} for more details. With this exact and smooth reformulation, standard nonlinear solvers can deal with the safety constraints.

\subsection{Objective function and overall optimization formulation for planning}
The objective is to minimize the deviation between the trajectory updated by the vehicle kinematic model and the reference trajectory associated with the optimal decision; meanwhile, smooth control actions are preferred for comfort. This objective function is defined as follows,
\begin{equation} \label{c0}
  \begin{split}
 J_p = ~ &  \sum_{i=0}^{M}  \left\| \mathbf{Q}_{\tau}(\tau[k+i|k] - \tau_{opt}^d[k+i]) \right\|_2^2  \\
 & + \sum_{i=0}^{M-1} \left\| \mathbf{Q}_{\mathbf{u}} \mathbf{u}[k+i] \right\|_2^2 \\
 & + \sum_{i=1}^{M-1} \left\| \mathbf{Q}_{\Delta\mathbf{u}} \Delta\mathbf{u}[k+i] \right\|_2^2 \\
  \end{split},
\end{equation}
where $\tau[k+i|k]$ represents the state at the time step $k+i$ predicted by the state at the time step $k$, $M$ is the length of optimization horizon under the sampling period $\Delta t$, $\mathbf{Q}_\tau$, $\mathbf{Q}_\mathbf{u}$ and $\mathbf{Q}_{\Delta \mathbf{u}}$ denote the corresponding weight matrices. The first term of the objective function penalizes the divergence of the solved trajectory from the reference trajectory. The second and third term suppress the excessive control actions and control input rates, respectively.

With the objective function and constraints introduced in the above subsections, the trajectory planning is formulated as an optimization problem, i.e.,
\begin{equation}\label{overall-opti}
\begin{aligned}
  \min_{\mathbf{u}, \mathbf{\lambda}, \mathbf{\mu}, \rho} & ~~ J_p \\
  \mbox{s.t.} ~~ & \tau[k|k] = \tau_{opt}^d[k], \\
                 & (\ref{vkm}), (\ref{pc}), (\ref{sc-re}).
\end{aligned}
\end{equation}
where the first constraint guarantees the initial solved state $\tau[k|k]$ to be equal to the reference state $\tau_{opt}^d[k]$.

\subsection{Ipopt-based approach for trajectory planning}
To solve the optimization problem (\ref{overall-opti}), we propose an algorithm to plan the trajectory for a lane change based on interior point optimizer (Ipopt) \cite{wachter2006implementation}. Ipopt is a widely used numerical solver for large-scale nonlinear optimization problems.

The algorithm for trajectory planning based on Ipopt is described in Algorithm \ref{c4-ipopt}.
The optimization problem at the time step $k$ is established as (\ref{overall-opti}).
By solving this problem, a sequence of optimal control actions $\mathbf{U}^*[k]=\{\mathbf{u}^*[k], \cdots, \mathbf{u}^*[k+M-1]\}$ can be obtained. As the basic MPC algorithm in \cite{camacho2013model}, we only apply the first control action $\mathbf{u}^*[k]$ to the ego vehicle to update its state.
At the next time step ${k+1}$, the horizon moves forward and a new optimization problem is formulated and solved by our algorithm.
By repeating this procedure, the ego vehicle can obtain an optimal trajectory that is smooth, collision-free, and satisfies the vehicle kinematic model.
Till now we have formulated an optimization and an Ipopt-based algorithm to complete the task of trajectory planning.

\begin{algorithm}
  \caption{Ipopt-based approach for trajectory planning}
  \label{c4-ipopt}
  \small
  \SetAlgoLined
  \KwData{Reference trajectory $\tau_{opt}^d$, interactive trajectory prediction $\{\tau_{1,opt}^f, \cdots, \tau_{n,opt}^f, \cdots, \tau_{N,opt}^f\}$, ego vehicle's current state $\tau_{ego}[O]$}
  \KwResult{Planned trajectory $\tau_{ego}^p$}
  Initialization: $k = O$\;
  \While{$k \leq O + P - M$}{
      Reset variables, and initial state constraint\;
      \For{$m = 1 $ {\rm to} $ M$}{
        Reset predictive model (\ref{vkm})\;
        Reset mechanical and physical constraints (\ref{pc})\;
        Calculate ego vehicle's $\mathbf{A}$ and $\mathbf{b}$ using (\ref{occupied-A-b})\;
        \For{$n = 1 $ {\rm to} $N$}{
            Calculate $\mathbf{A}$ and $\mathbf{b}$ using $\tau_n^f[k+m]$ and (\ref{occupied-A-b})\;
            Reset safety constraints (\ref{sc-re})\;
        }
      }
      Reset the objective function (\ref{c0})\;
      Solve above optimization using Ipopt\;
      Update ego vehicle's state using (\ref{vkm})\;
      $k = k + 1$\;
  }
\end{algorithm}

\section{Simulation Study}
In this section, we show the effectiveness and flexibility of the proposed lane change method through simulations in different complex traffic scenarios and comparison with other methods.
The effectiveness is further discussed by analyzing the prediction of other vehicles’ interactive trajectories.

\subsection{Simulation environment}
An open-source automotive driving simulator \cite{sisl} is used to implement the simulation scenarios.
We consider two-lane or three-lane highway. All vehicles have the same shapes; i.e., 4 meters in length and 1.8 meters in width.
Only the ego vehicle uses our proposed method for lane change.
For other vehicles, the longitudinal motions and lateral lane change behaviors are governed by the Intelligent Driver Model (IDM) \cite{treiber2000congested} and the Minimizing Overall Braking Induced by Lane changes (MOBIL) \cite{kesting2007general} model, respectively.
In order to increase the complexity of the simulated scenarios, the parameters in the IDM and the MOBIL are randomly sampled from uniform distributions. The sampling period $\Delta t$ is set to be equal to 0.1s.

\subsection{Baselines}
In order to further demonstrate the superiority of our proposed method, we compare it with the following baselines.
\begin{itemize}
\item IDM+MOBIL: This baseline controls the lateral and longitudinal motions with MOBIL and IDM, respectively, which is the same as other vehicles.
\item Our proposed method without interactive trajectory prediction (denoted as Proposed W/O IP): This baseline is the same as our proposed method except that it uses constant velocity, instead of the interactive trajectory prediction method in Section \ref{itp}, to predict other vehicles’ trajectories.
\end{itemize}

\subsection{Quantitative analysis}
In this subsection, we make quantitative analysis for the proposed method. We totally conduct 100 simulations under different traffic densities. In each simulation, vehicles run on a three-lane highway, and the parameters and initial positions for each vehicle are random. The mean velocities of the ego vehicle and surrounding vehicles are listed in Table \ref{quantitative}.

We can see that ego vehicles equipped with our proposed method achieve the fastest speeds, demonstrating the effectiveness of interactive trajectory prediction.
Meanwhile the mean speed of other vehicles are faster than that of other baselines. This is because the ego vehicle also considers the driving efficiencies of surrounding vehicles when making decisions on lane changes. Therefore, their driving efficiencies will be improved when the ego vehicle cooperates with others.
The ego vehicle using IDM+MOBIL only considers the observation at the current time but ignores the prediction of the future, so the mean speeds are the slowest.

\begin{table}
\caption{Mean speeds of the ego vehicle and other vehicles under 100 simulations.}
\label{quantitative}
\centering
\begin{center}
\begin{tabular}{p{70pt}<{\centering}p{40pt}<{\centering}p{35pt}<{\centering}p{35pt}<{\centering}}
\hline
\hline
 \multirow{2}{*}{Object}& \multirow{2}{*}{IDM+MOBIL} & Proposed & \multirow{2}{*}{Proposed} \\
  & &  W/O IP &  \\[1pt]
\hline
 Ego vehicle (m/s) & 13.95 & 14.85 & \textbf{16.01} \\[1pt]
\hline
 Ohter vehicles (m/s) & 14.19 & 14.31 & \textbf{14.56} \\[1pt]
\hline
\hline
\end{tabular}
\end{center}
\end{table}

\subsection{Case study}

\begin{figure}
\centering
\subfigure[Obstacle avoidance in dense traffic.]{
\includegraphics[width=8cm]{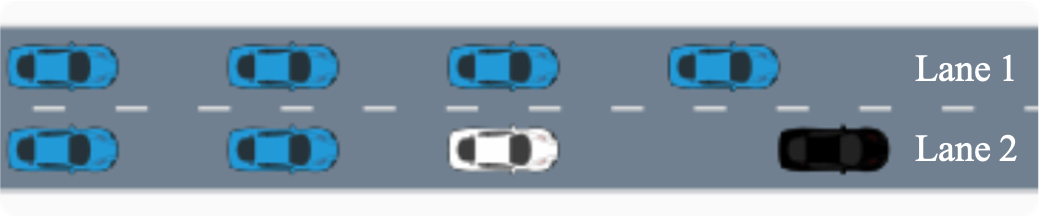}
}
\quad
\subfigure[Complex traffic with three lanes having different speeds and space gaps.]{
\includegraphics[width=8cm,height=1.8cm]{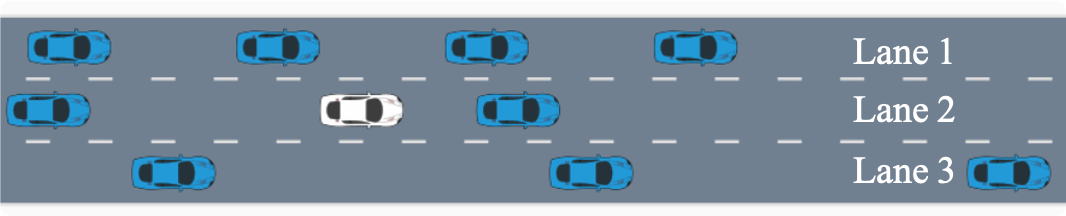}
}
\caption{Top views of the studied traffic scenarios. The white vehicle is the ego vehicle, the black one represents a stationary obstacle, and the blue vehicles are normal.}
\label{case}
\end{figure}

As shown in Fig. \ref{case}, we also study two practical cases for lane change, including obstacle avoidance in dense traffic, and complex traffic with three lanes having different speeds and space gaps.

\subsubsection{Case 1}
In this case, as shown in Fig. \ref{case}(a), the ego vehicle drives in dense traffic, and the space gap between any two consecutive vehicles in Lane 1 is insufficient for others to merge into. There is a stationary obstacle in front of the ego vehicle, so the ego vehicle should decide whether to stay in the current lane and stop or change to another lane for normal driving. If the ego vehicle chooses the latter decision for a higher efficiency, it needs the ability to interact with other vehicles to make them yield and make a space for it to cut in.

\begin{figure}
\centering
\subfigure[Time instant $t$ = 0s.]{
\includegraphics[width=9cm]{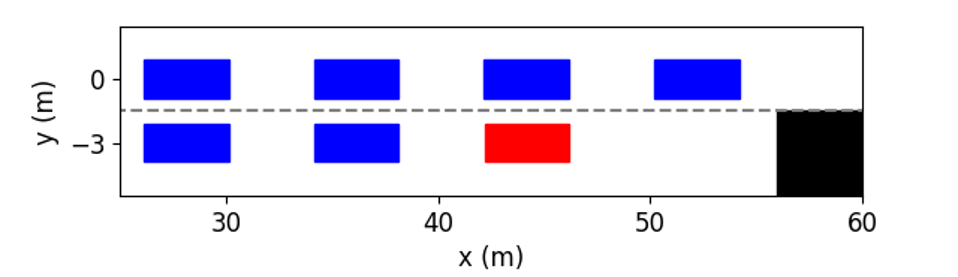}
}
\quad
\subfigure[Time instant $t$ = 2s.]{
\includegraphics[width=9cm]{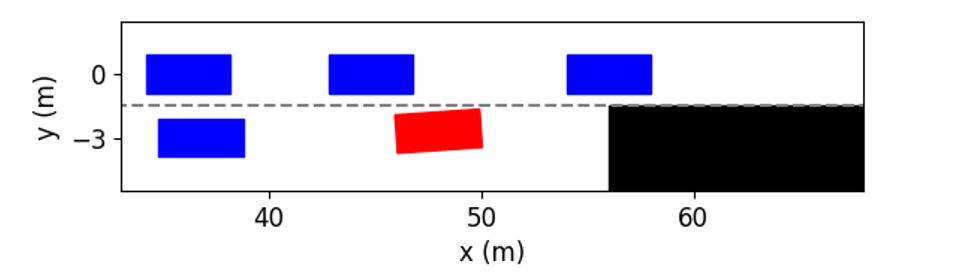}
}
\quad
\subfigure[Time instant $t$ = 3s.]{
\includegraphics[width=9cm]{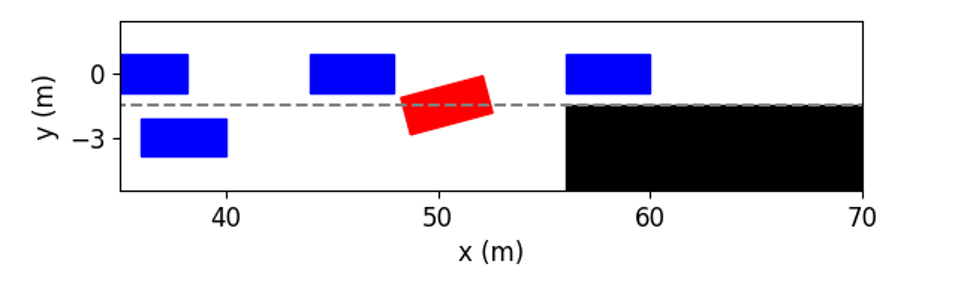}
}
\quad
\subfigure[Time instant $t$ = 6s.]{
\includegraphics[width=9cm]{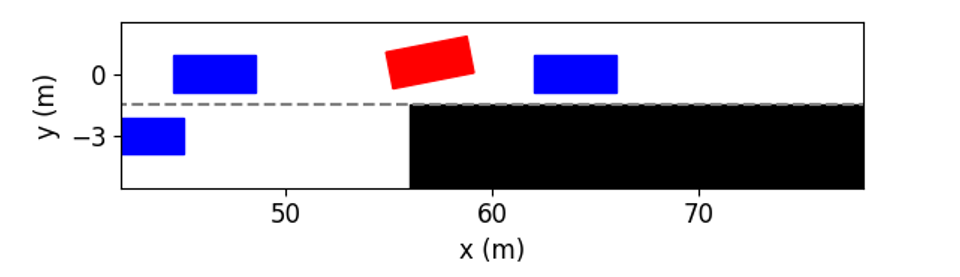}
}
\quad
\subfigure[Time instant $t$ = 8s.]{
\includegraphics[width=9cm]{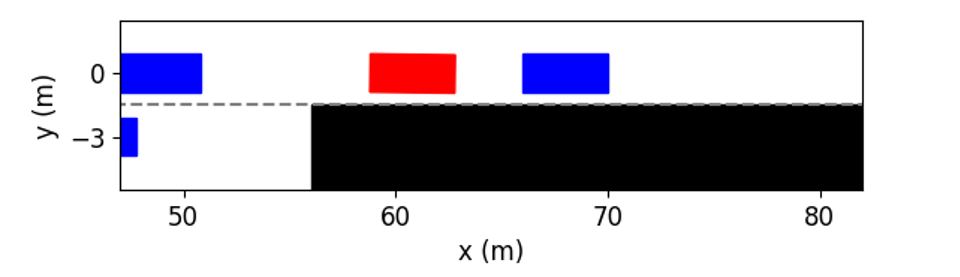}
}
\caption{Visualization of the testing results based on the proposed lane change method in the first case. Rectangles in red, blue, and black represent the ego vehicle, other vehicles, and the stationary obstacle, respectively. The gray dashed line indicates the lane boundary.}
\label{case1-t}
\end{figure}

Fig. \ref{case1-t} shows the visualization of testing results at different time instants.
It can be seen that although the initial inter-vehicle distances in Lane 1 are too narrow to merge in, the ego vehicle still successfully changes to that lane.
As shown in Fig. \ref{case1-t}(b), at the time instant $t$ = 2s, the ego vehicle heads to the left and is going to overtake the left vehicle, which slows down to help the ego vehicle cut in.
As shown in Fig. \ref{case1-t}(c), at the time instant $t$ = 3s, the ego vehicle has overtaken the left vehicle and is merging into the target lane.
Finally, as shown in Fig. \ref{case1-t}(e), the ego vehicle successfully changes to Lane 1 at about 8s.

\begin{figure}
\centerline{\includegraphics[scale=0.34]{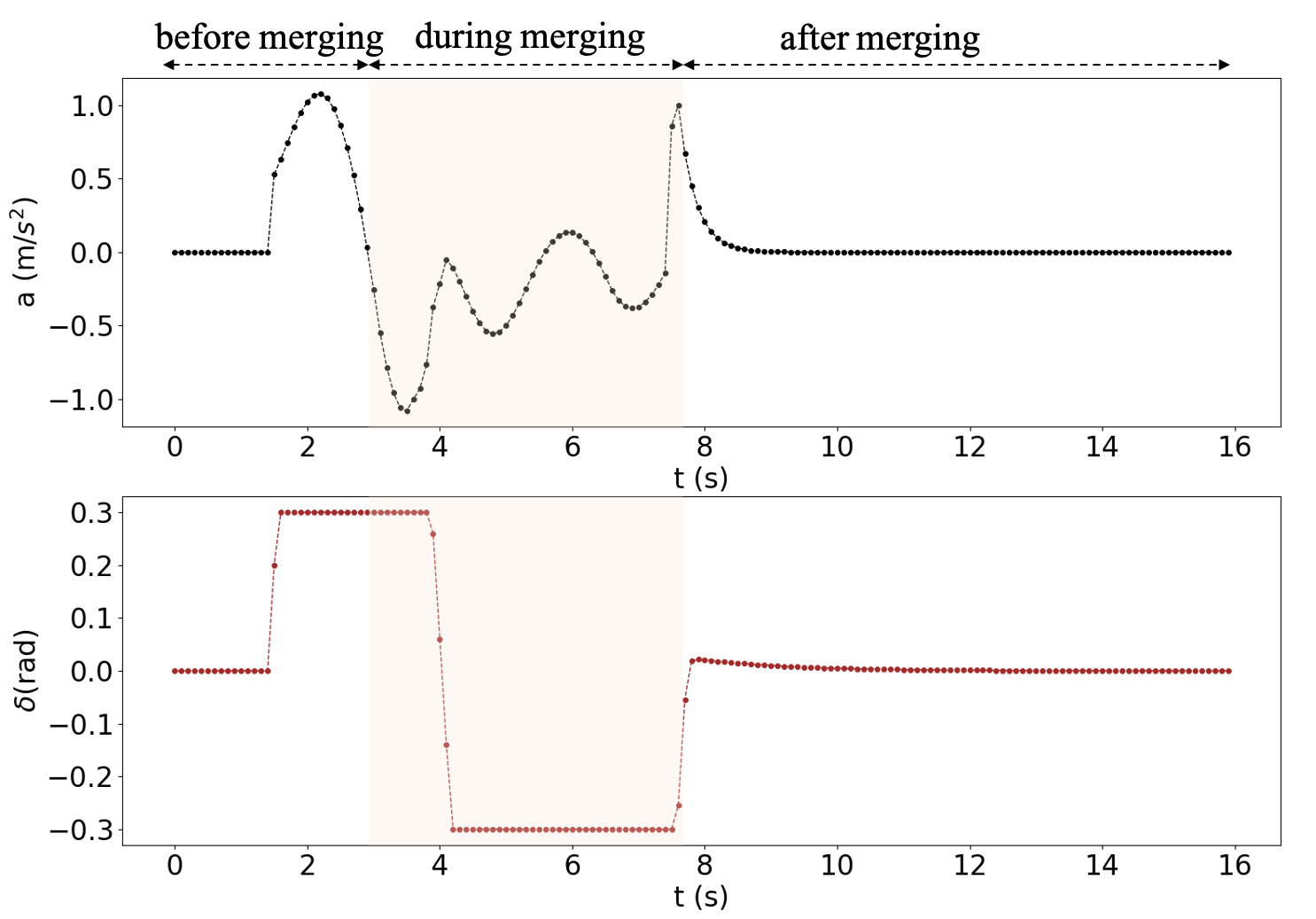}}
\caption{The acceleration and steering angle of the ego vehicle in the first case.}
\label{case1-control}
\end{figure}

Fig. \ref{case1-control} illustrates how the ego vehicle achieves interactions with surrounding vehicles by adjusting its acceleration and steering angle.
The whole process for lane change can be divided into three stages: before, during, and after merging.
Between 1$\sim$3s, it is before the merging. The ego vehicle prepares for lane change by applying positive steering angle to approach the target lane and accelerating to overtake the left vehicle.
When the ego vehicle is close enough to the target lane, as shown in Fig. \ref{case1-t}(b), the gap in front of the left vehicle is larger than the initial one, demonstrating the left vehicle is slowing down and giving way to the ego vehicle.
Between 3$\sim$7.8s, it is the second stage: the ego vehicle frequently adjusts its velocity and heading angle.
This is because ego vehicle needs to speed up to overtake the left vehicle to merge into Lane 1, but must slow down to avoid collision whenever it is too close to the preceding vehicle.
In the third stage, the ego vehicle has successfully entered the target lane.
At this time, the ego vehicle only keeps the lane and does not need to interact with others to seek cooperation, so its acceleration and steering angle gradually tend to zeros.

\begin{figure}
\centerline{\includegraphics[scale=0.35]{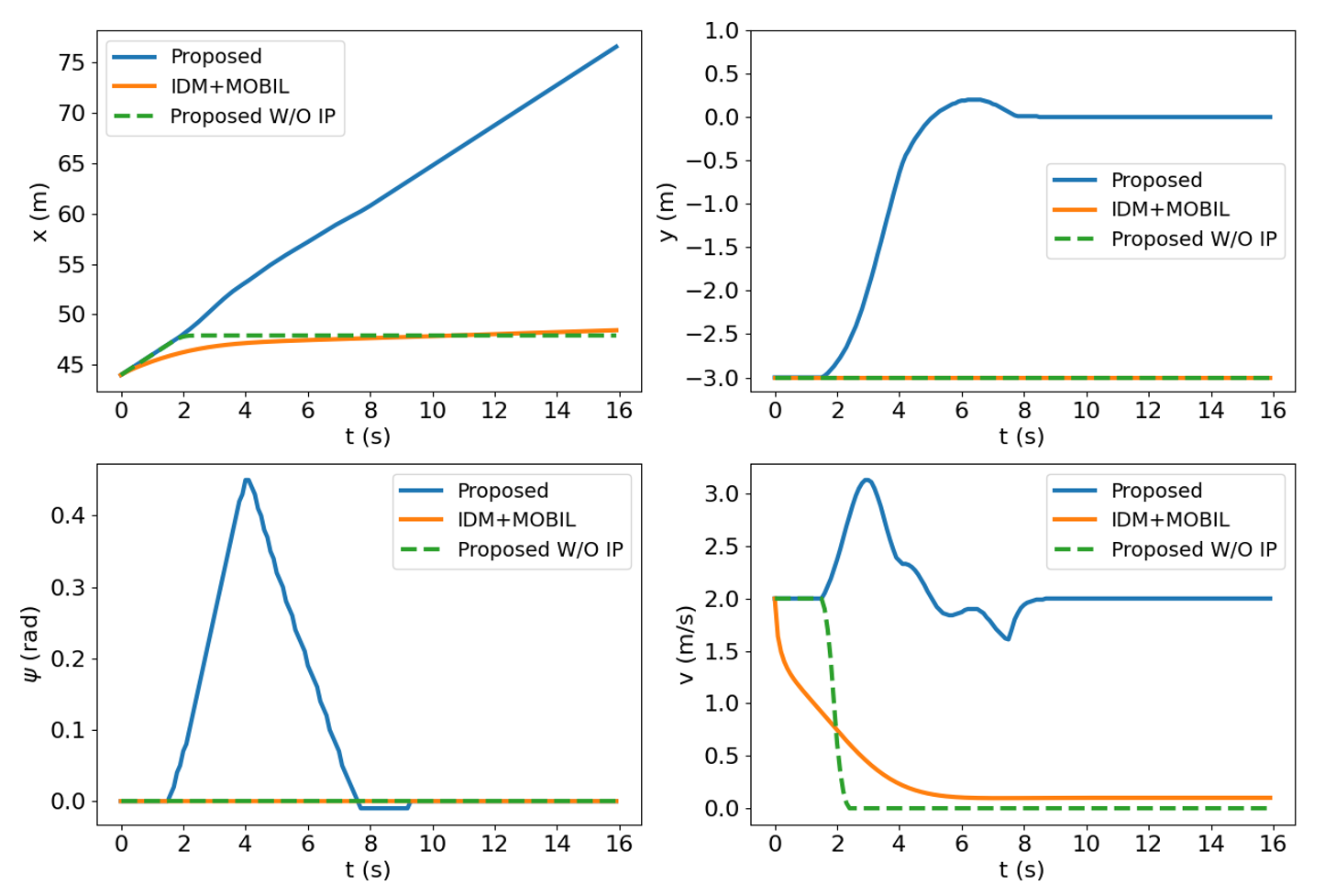}}
\caption{Comparison of vehicle states generated by the proposed lane change method and baselines in the first case.}
\label{case1-compare}
\end{figure}

To demonstrate the superiority of our proposed method, the comparison between our method and the above baselines is shown in Fig. \ref{case1-compare}.
We can see that both baselines choose a relatively conservative strategy. They decide to stay in the current lane, and thus the ego vehicle has to slow down to avoid collision with the obstacle.
This is because these baselines cannot utilize the cooperative behaviors of surrounding vehicles to help the ego vehicle change to another lane.
As a result, even if a higher speed can be gained by changing lanes, the ego vehicle is not advised to change lanes.
In contrast, our proposed method can make the ego vehicle actively interact with other vehicles, and thus the ego vehicle changes to Lane 1 safely to avoid being blocked by the obstacle and keep a high speed.

\begin{figure}
\centering
\subfigure[Left lane change and keeping in the current speed range.]{
\includegraphics[width=8cm]{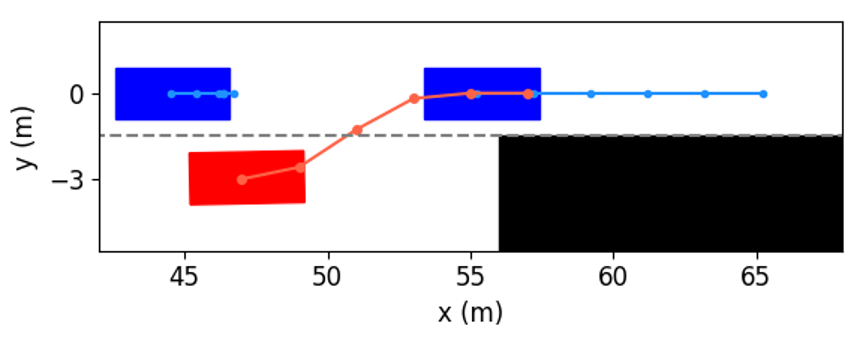}
}
\quad
\subfigure[Left lane change and accelerating.]{
\includegraphics[width=8cm]{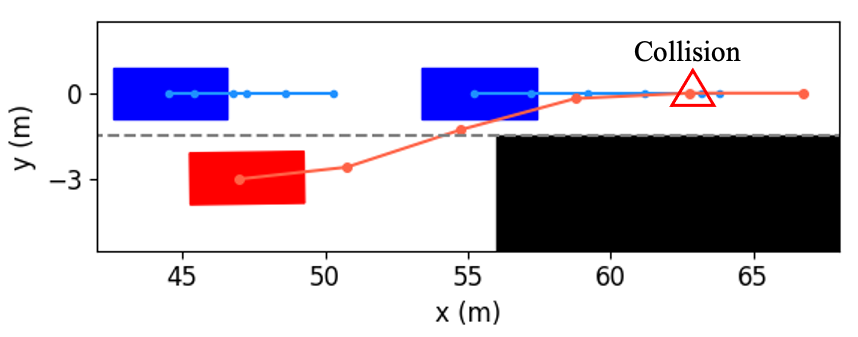}
}
\quad
\subfigure[Staying in the current lane and speed range.]{
\includegraphics[width=8cm]{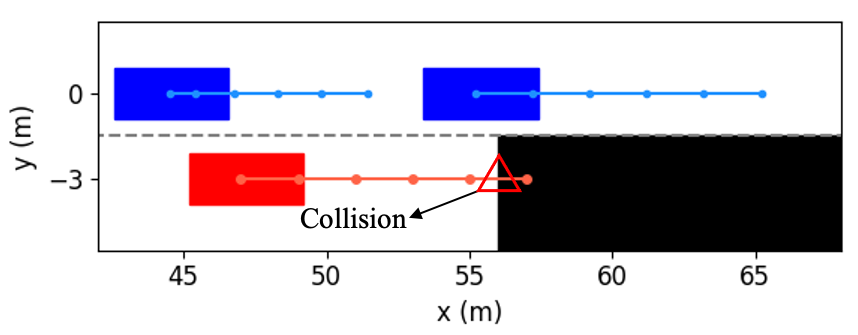}
}
\quad
\subfigure[Staying in the current lane and accelerating.]{
\includegraphics[width=8cm]{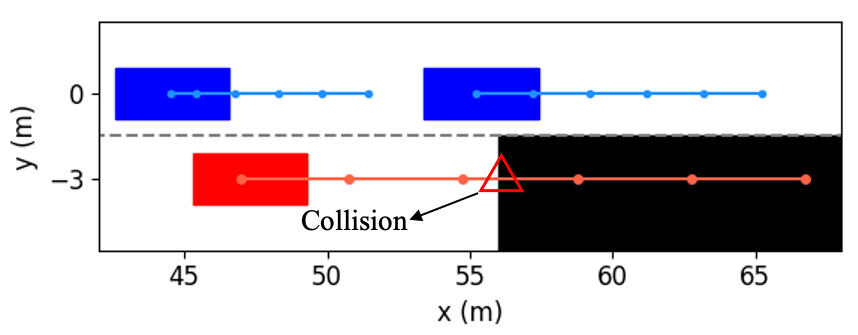}
}
\caption{Interactive trajectory prediction under different decisions in the first case. The blue lines represent the prediction, and the red lines are the reference trajectory generated by the method in Section \ref{generate}.}
\label{case1-p}
\end{figure}

In order to further investigate how our method determines whether the lane change is a feasible decision, and how to use vehicular interactions to achieve cooperation during trajectory planning, Fig. \ref{case1-p} shows several examples of the interactive trajectory prediction in this case.
Our proposed method selects the decision in Fig. \ref{case1-p}(a) as the optimal one, since the interactive trajectory prediction indicates that if the ego vehicle runs in this way, the vehicle behind will slow down and enlarge the gap.
Therefore, this decision can help the ego vehicle avoid collision with the obstacle and change lanes safely to ensure normal driving.
The decision in Fig. \ref{case1-p}(b) can also make the ego vehicle overtake the left vehicle and then change lanes, but the prediction shows that the ego vehicle would collide with the front vehicle.
This is consistent with the intuition that the front vehicle rarely reacts to the speed of the following vehicle.
We can see from Fig. \ref{case1-p}(c) and (d) that if the ego vehicle neither changes to another lane nor decelerates, it will collide with obstacles.
Based on the above discussion and analysis, we conclude that our proposed method is able to obtain reliable interactive trajectory prediction, which can be further used to determine the most proper decision and provide the motion planning module with safety constraints to avoid collision between vehicles.

\subsubsection{Case 2}
In this case, a complex traffic scenario is studied.
Fig. \ref{case}(b) shows the top view snapshot of vehicles at the initial time.
The vehicles in Lane 1 have the fastest speeds and narrowest inter-vehicle distances, so it is risky to change to this lane in order to obtain a higher speed.
On the contrary, vehicles in Lane 3 have the slowest speeds and largest inter-vehicle distances.
The ego vehicle is driving in Lane 2, and vehicles in Lane 2 have moderate speeds. Since the vehicle in front of the ego vehicle travels at a slower speed than vehicles in Lane 1, the ego vehicle needs to decide whether to follow the preceding vehicle and keep a slow speed to guarantee the safety or to change to Lane 1 to gain a higher speed. The decision is dependent on the reaction of surrounding vehicles and whether a feasible trajectory can be generated.

\begin{figure*}
\centering
\subfigure[Time instant $t$ = 0s.]{
\includegraphics[width=5.4cm]{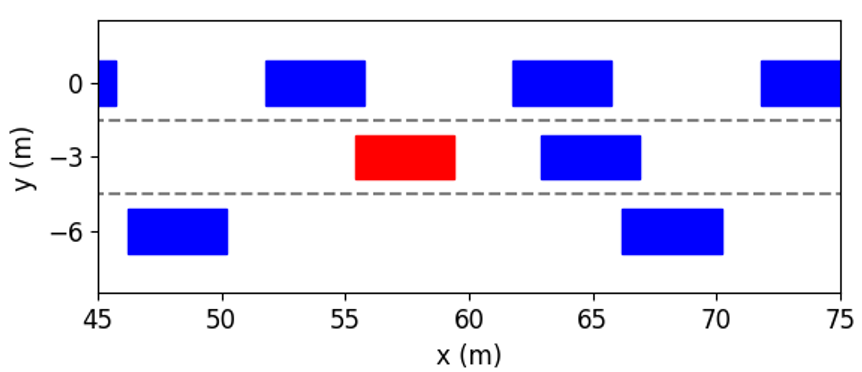}
}
\quad
\subfigure[Time instant $t$ = 2s.]{
\includegraphics[width=5.4cm]{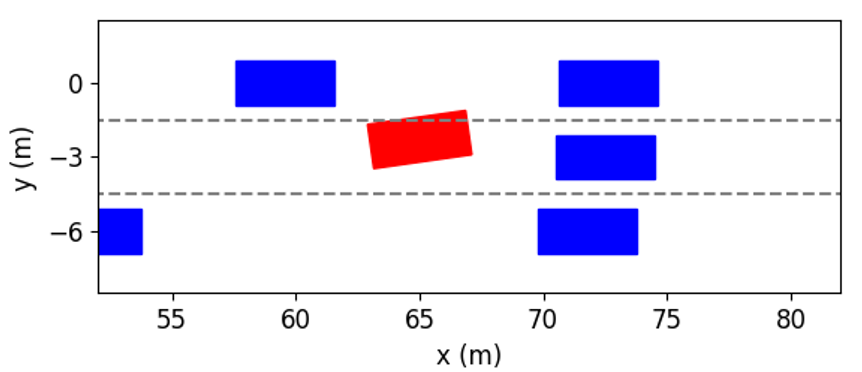}
}
\quad
\subfigure[Time instant $t$ = 4s.]{
\includegraphics[width=5.4cm]{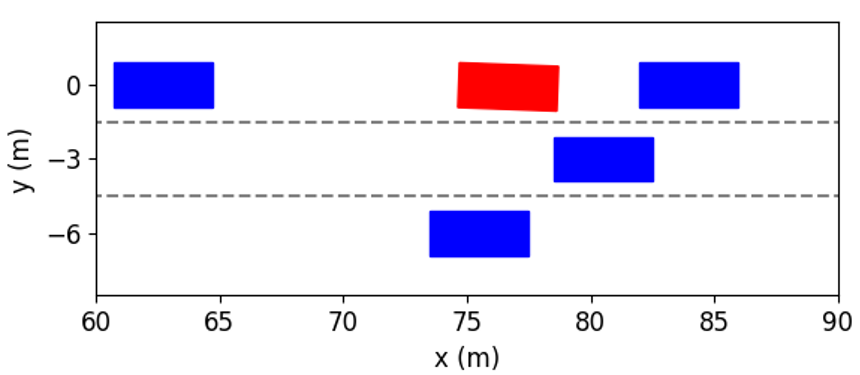}
}
\quad
\subfigure[Time instant $t$ = 7s.]{
\includegraphics[width=5.4cm]{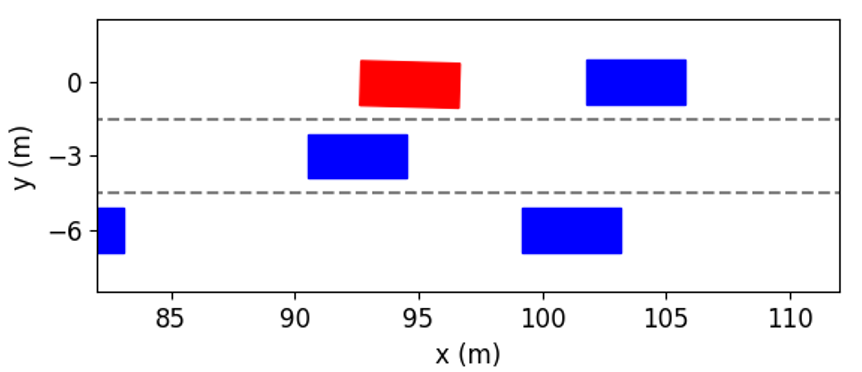}
}
\quad
\subfigure[Time instant $t$ = 7.5s.]{
\includegraphics[width=5.4cm]{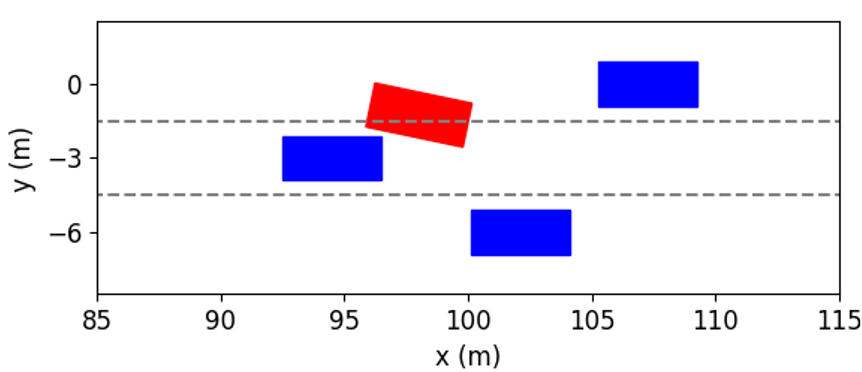}
}
\quad
\subfigure[Time instant $t$ = 10s.]{
\includegraphics[width=5.4cm]{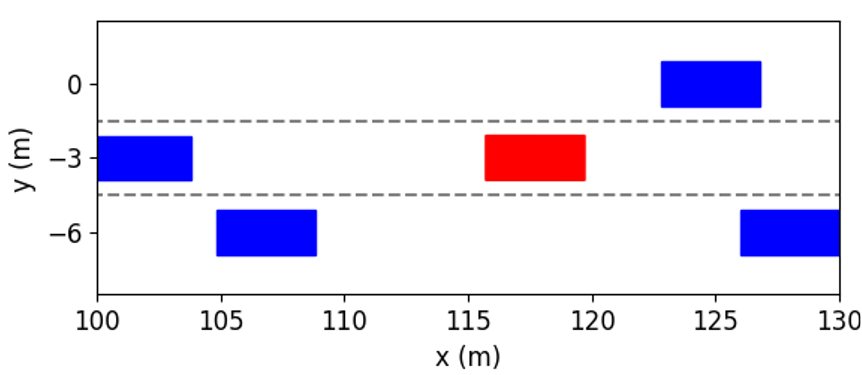}
}
\caption{Visualization of the testing results based on the proposed lane change method in the second case. }
\label{case2-t}
\end{figure*}

The simulation results at different time instants are presented in Fig. \ref{case2-t}. We can see that the ego vehicle makes two lane changes to reach the fastest speed.
At first, in order to get rid of the influence of the preceding vehicle with slow speeds, the ego vehicle tries to change to Lane 1 by interacting with other vehicles.
As shown in Fig. \ref{case2-t}(b), at the time instant $t$ = 2s, the left following vehicle slows down to expand the space gap to help the ego vehicle to cut in.
As shown in Fig. \ref{case2-t}(c), the ego vehicle completes lane change and overtaking at the time instant $t$ = 4s.
Further, as shown in Fig. \ref{case2-t}(f), the ego vehicle returns to Lane 2 and obtains the fastest speed.

\begin{figure}
\centerline{\includegraphics[scale=0.33]{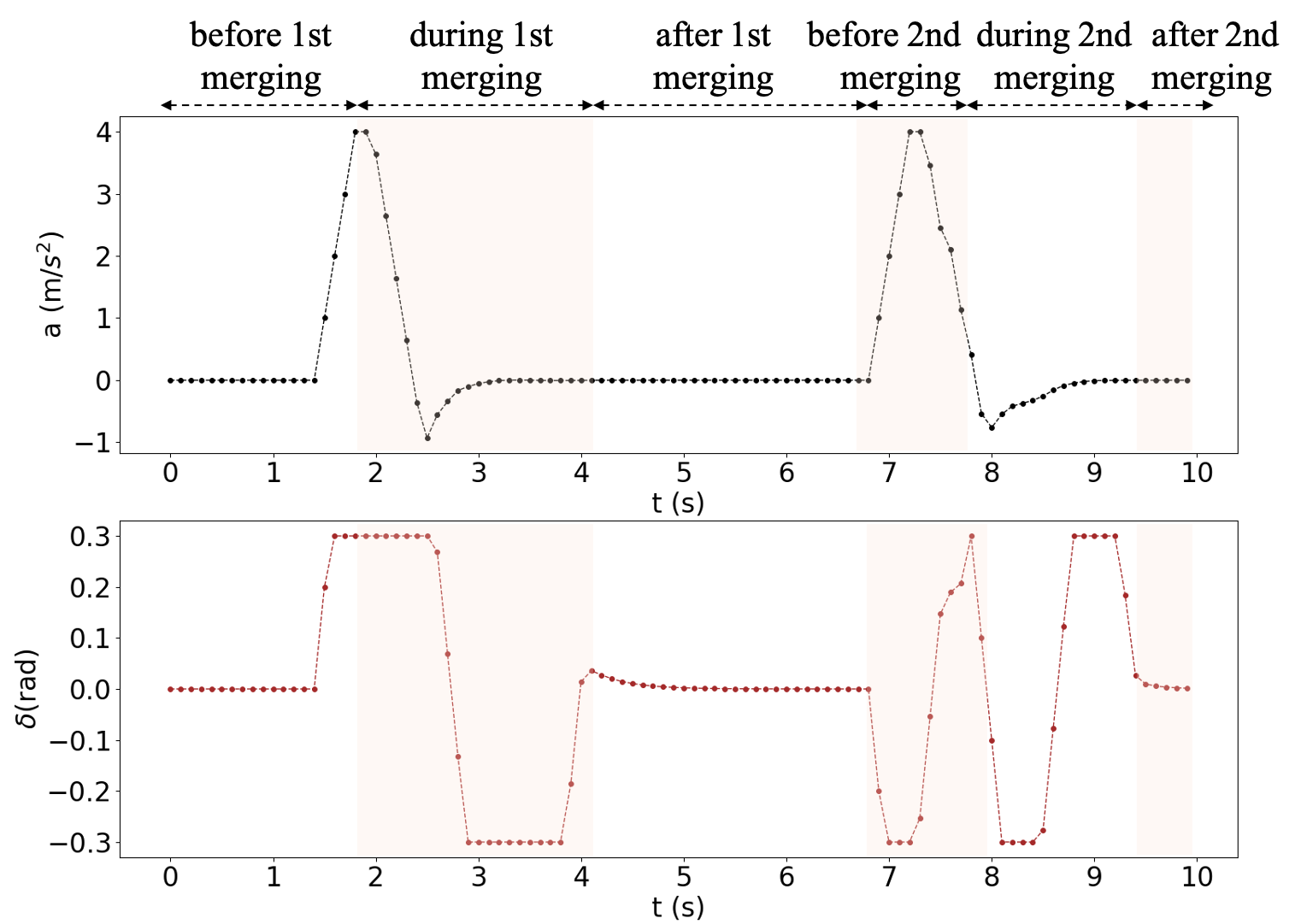}}
\caption{The acceleration and steering angle of the ego vehicle in the second case.}
\label{case2-control}
\end{figure}

These interactive behaviors can also be interpreted by the acceleration and steering angle of the ego vehicle in Fig. \ref{case2-control}. Similar to the first case, the process for two lane changes is divided into six stages.
Since the initial speed of the ego vehicle is lower than the left vehicle’s, it increases its longitudinal speed while moving towards Lane 1 before the first merging.
At about 2 to 4 seconds, the acceleration and steering angle of the ego vehicle fluctuate, which indicates that the ego vehicle wants to switch to the target lane quickly while avoiding collision with the preceding vehicle.
Between 4$\sim$6.7s, the control actions are getting closer to zeros since the ego vehicle has finished the first lane change and only needs to drive steadily.
At about 7 seconds, as shown in Fig. \ref{case2-t}(d), the space gap between the front vehicle and the right one is large enough, and there is no front vehicle in the Lane 2.
Consequently, after the prediction and evaluation, our proposed method decides to accelerate and change to Lane 2.
Between 8$\sim$9.5s, the ego vehicle smoothly enters the target lane.

\begin{figure}
\centerline{\includegraphics[scale=0.35]{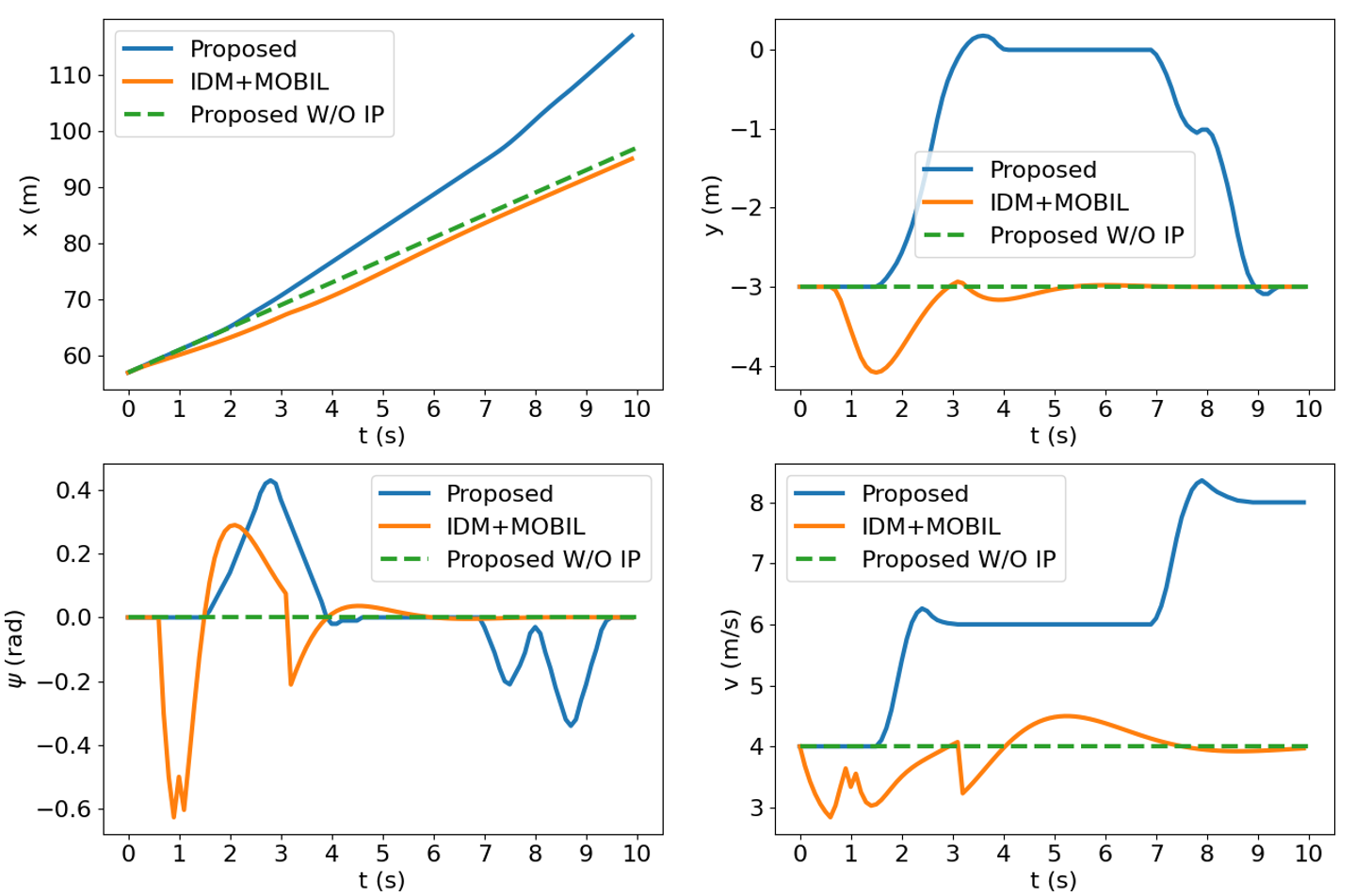}}
\caption{Comparison of vehicle states generated by the proposed lane change method and baselines in the second case.}
\label{case2-compare}
\end{figure}

As shown in Fig. \ref{case2-compare}, in the second studied case, the ego vehicle using both baselines exhibits conservative behaviors. For the baseline Proposed W/O IP, we can see that it decides to stay in the current lane and speed. Because of the lack of interactive prediction, this baseline believes there would be a collision when changing to Lane 1. For another baseline, we can see that at about 0.5 seconds, the ego vehicle attempts to switch to Lane 3 where the space gap between vehicles is large.
But soon it finds that the speeds of vehicles in this lane are too slow, so it returns to the original lane.
This is because this baseline cannot predict the future trajectory of other vehicles.
Therefore, when it decides to change to Lane 3, it only considers the benefits of a larger space gap, but cannot predict the future speed loss.
The lack of trajectory prediction leads to large amplitudes and frequent actions on the steering wheel, which reduces the driving comfort.
After returning to the original lane, the ego vehicle has to follow the preceding vehicle and maintain a relatively slow speed.
Similar to the first case, although the speeds of vehicles in Lane 1 are faster, the ego vehicle is not recommended to change that lane because the close distances between vehicles may cause a collision.
However, the ego vehicle equipped with our proposed method is more inclined to create a favorable environment through active cooperation with other vehicles, so it can achieve safe, comfortable and efficient driving.

\begin{figure*}
\centering
\subfigure[Left lane change and decelerating.]{
\includegraphics[width=5.4cm]{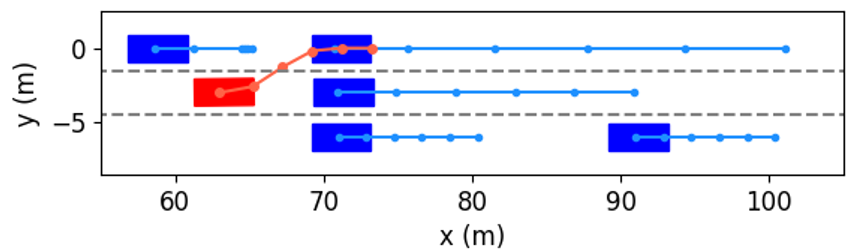}
}
\quad
\subfigure[Left lane change and keeping in the current speed range.]{
\includegraphics[width=5.4cm]{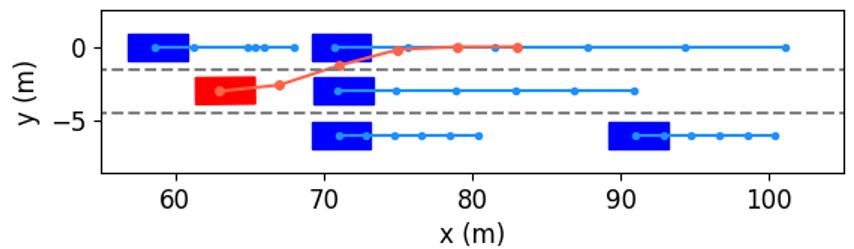}
}
\quad
\subfigure[Left lane change and accelerating.]{
\includegraphics[width=5.4cm]{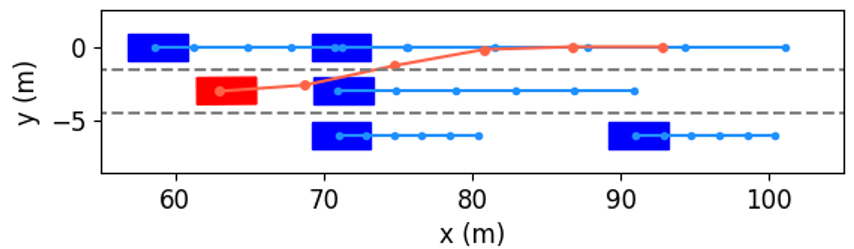}
}
\quad
\subfigure[Staying in the current lane and decelerating.]{
\includegraphics[width=5.4cm]{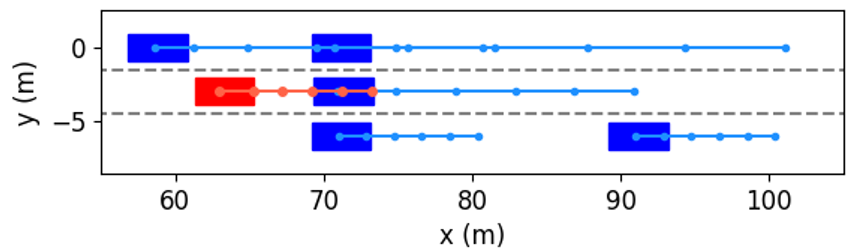}
}
\quad
\subfigure[Staying in the current lane and speed range.]{
\includegraphics[width=5.4cm]{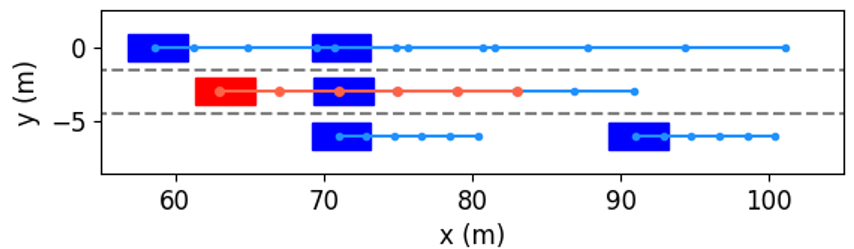}
}
\quad
\subfigure[Staying in the current lane and accelerating.]{
\includegraphics[width=5.4cm]{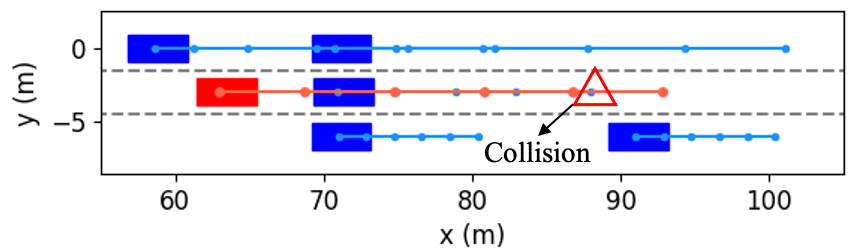}
}
\quad
\subfigure[Right lane change and decelerating.]{
\includegraphics[width=5.4cm]{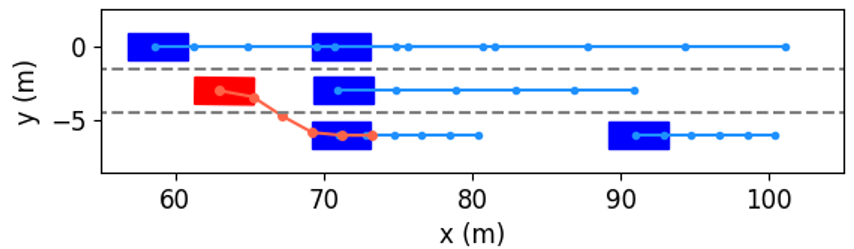}
}
\quad
\subfigure[Right lane change and keeping in the current speed range]{
\includegraphics[width=5.4cm]{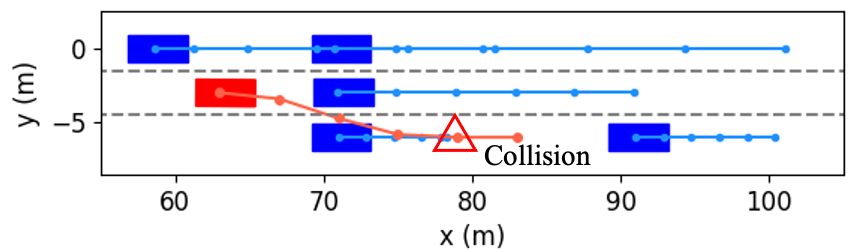}
}
\quad
\subfigure[Right lane change and accelerating.]{
\includegraphics[width=5.4cm]{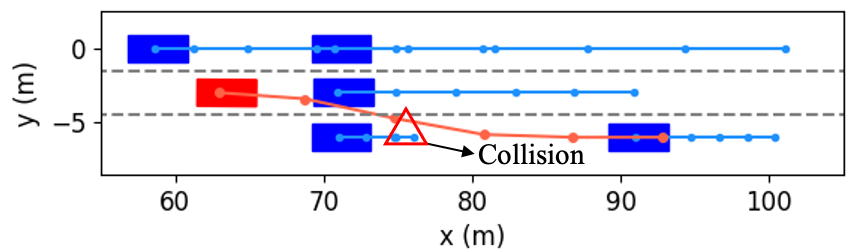}
}
\caption{Interactive trajectory prediction under different candidate decisions in the first decision process of the second case. }
\label{case2-p1}
\end{figure*}

We take the first merging as an example to explore how the proposed method decides to accelerate and change to Lane 1.
The interactive trajectory prediction for each decision is shown in Fig. \ref{case2-p1}, and the cost values are listed in Table \ref{case2-cost}.
From the prediction in Fig. \ref{case2-p1}(f), (h) and (i), we can see that these three decisions would lead to collisions, and thus their cost values for safety are much higher than others’.
When the ego vehicle changes to the left lane at low or current speeds, the prediction in Fig. \ref{case2-p1}(a) and (b) indicate that the left following vehicle will slow down.
Despite both decisions are safe, their cost values for efficiency vary greatly, resulting in a large difference in the total costs.
Besides, although keeping in the current lane and speed range would not cause any costs for safety and comfort, its efficiency cost contributes to a higher total cost than that of left lane change \& accelerating.
Therefore, our proposed method selects left lane change \& accelerating as the optimal decision.
From the results in Fig. \ref{case2-t}$\sim$\ref{case2-compare}, we can see that the planned trajectory is safe and smooth, has low requirements for control actions, and achieves higher driving efficiency, demonstrating it is an appropriate decision.

\begin{table}
\caption{Cost values for different candidate decisions in the first decision process of the second case.}
\label{case2-cost}
\centering
\begin{center}
\begin{tabular}{p{60pt}<{\centering}|p{20pt}<{\centering}p{21pt}<{\centering}p{30pt}<{\centering}|p{25pt}<{\centering}}
\hline
\hline
\multirow{2}{*}{Decision} & \multirow{2}{*}{Safety} & \multirow{2}{*}{Comfort} & \multirow{2}{*}{Efficiency} & Total\\
 & & & & cost \\
\hline
Left lane change   & \multirow{2}{*}{39.6} & \multirow{2}{*}{388.3}  & \multirow{2}{*}{8094.8} & \multirow{2}{*}{8522.7} \\
\& decelerating &  &   &  &   \\[1pt]
Left lane change & \multirow{2}{*}{23.6} & \multirow{2}{*}{262.0}  & \multirow{2}{*}{5020.4} & \multirow{2}{*}{5306.0}  \\
\& speed keeping &  &   &  &   \\[1pt]
Left lane change   & \multirow{2}{*}{749.8} & \multirow{2}{*}{388.3}  & \multirow{2}{*}{\textbf{1633.2}} & \multirow{2}{*}{\textbf{2771.3}}  \\
\& accelerating &  &   &  &   \\[1pt]
Lane keeping & \multirow{2}{*}{290.0} & \multirow{2}{*}{126.3}  & \multirow{2}{*}{6293.1} & \multirow{2}{*}{6709.4}  \\
\& decelerating &  &   &  &   \\[1pt]
Lane keeping   & \multirow{2}{*}{\textbf{0.0}} & \multirow{2}{*}{\textbf{0.0}}  & \multirow{2}{*}{3600.0} & \multirow{2}{*}{3600.0}  \\
\& speed keeping &  &   &  &   \\[1pt]
Lane keeping & \multirow{2}{*}{8139.0} & \multirow{2}{*}{126.3}  & \multirow{2}{*}{1669.1} & \multirow{2}{*}{9934.4}  \\
\& accelerating &  &   &  &   \\[1pt]
Right lane change   & \multirow{2}{*}{211.8} & \multirow{2}{*}{388.3}  & \multirow{2}{*}{6094.8} & \multirow{2}{*}{6694.9}  \\
\& decelerating &  &   &  &   \\[1pt]
Right lane change & \multirow{2}{*}{7468.0} & \multirow{2}{*}{262.0}  & \multirow{2}{*}{3520.4} & \multirow{2}{*}{11250.4}  \\
\& speed keeping &  &   &  &   \\[1pt]
Right lane change   & \multirow{2}{*}{3051.3} & \multirow{2}{*}{388.3}  & \multirow{2}{*}{1633.2} & \multirow{2}{*}{5072.8}  \\
\& accelerating &  &   &  &   \\[1pt]
\hline
\hline
\end{tabular}
\end{center}
\end{table}

\section{Conclusions}
In this paper, we have presented a cooperation-aware method that can help the ego vehicle to execute a proper lane change for guarantee driving efficiency and safety.
Different from the existing methods, we have considered the interdependence between the lane change decision module and the trajectory planning module so as to avoid possible conflicts between them.
An interactive trajectory prediction method has been proposed, with which we have designed a candidate evaluation method.
By doing so, our approach can find potential cooperation by predicting others’ interactions so that a less conservative decision on lane change can be made. With a right lane change decision, a safe, efficient, and comfortable driving can be well guaranteed.
During the process of planning a lane change trajectory, interactive trajectories has been incorporated into constraints, such that the ego vehicle can interact with others to promote cooperation by adjusting its control inputs.
The quantitative results have demonstrated that when using our proposed method, the driving efficiencies of the ego vehicle and others are 14.8\% and 2.6\% higher than those with methods without interactive prediction.
Two practical case studies further analyze how the proposed method predicts interactive trajectories of other vehicles and uses this interactive prediction to make decisions.

Following the stream of this study, several future works can be done. (i) The uncertainty caused by sensor measurement errors and communication delays can be considered to improve the robustness. (ii) The proposed lane change method is only applied on single vehicle to improve driving efficiencies of its nearby vehicles and itself. Based on our method, exploring a distributed framework to optimize a large traffic network is also worthwhile for future research.



\ifCLASSOPTIONcaptionsoff
  \newpage
\fi

%
{\small
\bibliographystyle{IEEEtran}
\bibliography{IEEEabrv,egbib}
}

%






\end{document}